\documentclass[letterpaper, 10 pt, conference]{ieeeconf}

\usepackage{silence}
\usepackage{amsmath,amsfonts}
\usepackage{amssymb}
\usepackage{algorithm}
\usepackage{algorithmic}
\usepackage{array}
\usepackage{textcomp}
\usepackage{stfloats}
\usepackage{url}
\usepackage{verbatim}
\usepackage{graphicx}
\usepackage{balance}
\usepackage{svg}
\usepackage{booktabs}
\usepackage{multirow}
\usepackage{colortbl}
\usepackage{bm}
\usepackage{subcaption}
\usepackage{glossaries}

\makeatletter
\let\NAT@parse\undefined
\makeatother
\usepackage [hidelinks]{hyperref}
\usepackage{cleveref}

\usepackage{listings}
\def\vc{{\bm{c}}}

\def\vf{{\bm{f}}}

\def\vk{{\bm{k}}}

\def\vs{{\bm{s}}}

\def\vu{{\bm{u}}}

\def\vB{{\bm{B}}}

\def\vG{{\bm{G}}}

\def\vJ{{\bm{J}}}

\def\vmu{{\bm{\mu}}}

\def\vpsi{{\bm{\psi}}}

\newacronym{rl}{RL}{Reinforcement Learning}
\newacronym{drl}{DRL}{Deep Reinforcement Learning}
\newacronym{hrl}{HRL}{Hierarchical Reinforcement Learning}
\newacronym{saferl}{SafeRL}{Safe Reinforcement Learning}
\newacronym{avi}{AVI}{Approximate Value-Iteration}
\newacronym{api}{API}{Approximate Policy-Iteration}
\newacronym[plural=MDPs, firstplural=Markov Decision Processes (MDPs)]{mdp}{MDP}{Markov Decision Process}
\newacronym{cmdp}{CMDP}{Constrained Markov Decision Processes}
\newacronym{safeexp}{SafeExp}{Safe Exploration}
\newacronym{kl}{KL}{Kullback-Leibler Divergence}
\newacronym{gae}{GAE}{Generalized Advantage Estimation}
\newacronym{papi}{PAPI}{Projections for Approximate Policy Iteration}
\newacronym{her}{HER}{Hindsight Experience Replay}
\newacronym{ham}{HAM}{Hierarchy of Abstract Machines}
\newacronym{mom}{MOM}{Measure of Manipulability}
\newacronym{bo}{BO}{Bayesian Optimization}
\newacronym{hebo}{HEBO}{Heteroscedastic Evolutionary Bayesian Optimisation}
\newacronym{ucb}{UCB}{Upper Confidence Bound}
\newacronym{pi}{PI}{Probability of Improvement}
\newacronym{ei}{EI}{Expected Improvement}
\newacronym{nl}{NLP}{Nonlinear Programming}
\newacronym{lp}{LP}{Linear Programming}
\newacronym{qp}{QP}{Quadratic Programming}
\newacronym{aqp}{AQP}{Anchored Quadratic Programming}
\newacronym{ode}{ODE}{Ordinary Differential Equation}
\newacronym{atacom}{ATACOM}{Acting on the TAngent Space of the COnstraint Manifold}
\newacronym{datacom}{D-ATACOM}{Distributional ATACOM}
\newacronym{atacomdc}{ATACOM-DC}{ATACOM with Directional Constraints}
\newacronym{ivp}{IVP}{Initial Value Problem}
\newacronym{rref}{RREF}{Reduced Row Echlon Form}
\newacronym{rcef}{RCEF}{Reduced Column Echlon Form}
\newacronym{cpo}{CPO}{Constrained Policy Optimization}
\newacronym{trpo}{TRPO}{Trust Region Policy Optimization}
\newacronym{sac}{SAC}{Soft Actor-Critic}
\newacronym{rmp}{RMP}{Riemannian Motion Policies}
\newacronym{dnn}{DNN}{Deep Neural Networks}
\newacronym{sdf}{SDF}{Signed Distance Function}
\newacronym{redsdf}{ReDSDF}{Regularized Deep Signed Distance Fields}
\newacronym{apf}{APF}{Artificial Potential Fields}
\newacronym{hri}{HRI}{Human-Robot Interaction}
\newacronym{poi}{PoI}{Point of Interest}

\newacronym{mpc}{MPC}{Model Predictive Control}
\newacronym{dcs}{DCS}{Directly Controllable State}
\newacronym{dus}{DUS}{Directly Uncontrollable State}

\newacronym{cbf}{CBF}{Control Barrier Function}
\newacronym{iss}{ISS}{Input-to-State Stable}
\newacronym{codf}{CoDF}{Constraint Decay Function}
\newacronym{fvf}{FVF}{Feasibility Value Function}
\newacronym{kfvf}{k-FVF}{k-step Feasibility Value Function}

\newacronym{var}{VaR}{Value-at-Risk}
\newacronym{cvar}{CVaR}{Conditional Value-at-Risk}

\IEEEoverridecommandlockouts
\title{\LARGE \bf Directional Constraints for Efficient Exploration \\ in Safe Reinforcement Learning}

\author{Paolo Magliano$^{1}$, Puze Liu$^{2,3}$, Jan Peters$^{4,3,5}$, Davide Tateo$^{6,4,\dagger}$, Raffaello Camoriano$^{1,7,\dagger}$%
\thanks{This study was carried out within the FAIR - Future Artificial Intelligence Research and received funding from the European Union Next-GenerationEU (PIANO NAZIONALE DI RIPRESA E RESILIENZA (PNRR) – MISSIONE 4 COMPONENTE 2, INVESTIMENTO 1.3 – D.D. 1555 11/10/2022, PE00000013). This manuscript reflects only the authors’ views and opinions, neither the European Union nor the European Commission can be considered responsible for them.}
\thanks{$^{1}$Dipartimento di Automatica e Informatica, Politecnico di Torino, Turin, Italy. %
{\tt raffaello.camoriano@polito.it}}%
\thanks{$^{2}$Tongji University, Shanghai Research Institute for Intelligent Autonomous Systems.
        {\tt puze\_liu@tongji.edu.cn}}%
\thanks{$^{3}$German Research Center for AI (DFKI).}%
\thanks{$^{4}$Intelligent Autonomous Systems Group, TU Darmstadt, Germany. {\tt jan.peters@tu-darmstadt.de} $^{5}$Hessian.AI. }%
\thanks{$^{6}$Lund University, Sweden.
        {\tt davide.tateo@cs.lth.se}}%
\thanks{$^{7}$Istituto Italiano di Tecnologia, Genoa, Italy. $^\dagger$Co-last authors.}%
}

\begin{document}
\maketitle

\begin{abstract}
Reinforcement Learning has revolutionized the landscape of robotic research, allowing robust learning of complex robotic skills in simulation. However, real-world deployment in open-ended environments requires strong safety guarantees to prevent dangerous or harmful behaviors.
Safe Reinforcement Learning methods address this requirement by enforcing safety constraints. Nevertheless, learning under constraints often reduces learning speed and could lead to suboptimal task performance, as the agent must solve a more complex constrained optimization problem compared to unconstrained settings.
To tackle this issue, in this work, we propose an extension of the ATACOM framework, a state-of-the-art reliable safety layer that can be integrated with existing Reinforcement Learning algorithms to enforce constraints derived from prior knowledge of the system or learned directly from data.
Our proposed method, named ATACOM Directional Constraints (ATACOM-DC), significantly improves the safety-performance trade-off by introducing directional constraints that distinguish between actions approaching and moving away from constraint boundaries, activating constraint enforcement only when necessary.
We evaluate our method across a range of challenging robotic control tasks in simulation, analyzing both constraint-violation costs and achieved task performance. Code and additional material at 
\mbox{\url{https://atacom-dc.robot-learning.net}}.

\end{abstract}

\section{Introduction}
\label{sec:introduction}
In recent years, \gls{rl} has become the dominant technique for learning complex, dynamic robotic skills, both in the area of locomotion~\cite{rudin2022learning,zhuang2023robot} and manipulation~\cite{huang2023dynamic,lin2024twisting,li2025morphologically}. 
Most of the proposed approaches rely on the concept of Domain Randomization~\cite{tobin2017domain,rudin2022learning}. The key idea is to generate policies that can be robustly deployed in the real world, reacting to different environmental conditions by training on a family of simulators offline.
However, when we deploy policies in the real world, policy robustness may not be sufficient due to safety-critical requirements in many applications.
Indeed, while safety violations in simulation are not problematic, in the real world they may damage the robot, the environment, or harm people. Relying only on disturbance rejection capabilities from a random distribution of the environment may not be sufficient and definitely lacks sound theoretical guarantees on the safety of the system.

\begin{figure}[t]
  \centering
  \begin{subfigure}[t]{0.31\columnwidth}
    \centering
    \includegraphics[width=0.95\columnwidth]{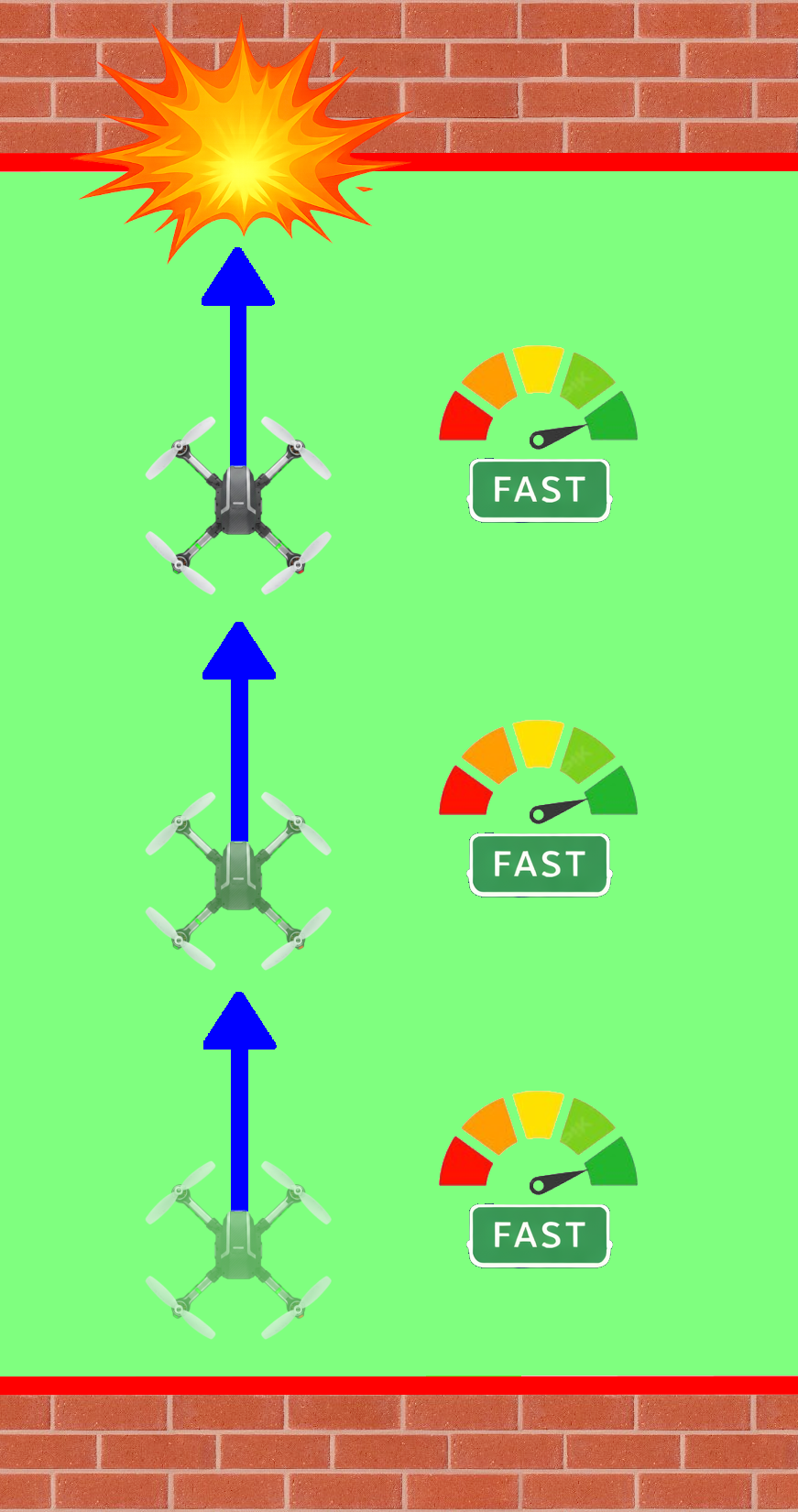}
  \end{subfigure}\hfill
  \begin{subfigure}[t]{0.31\columnwidth}
    \centering
    \includegraphics[width=0.95\columnwidth]{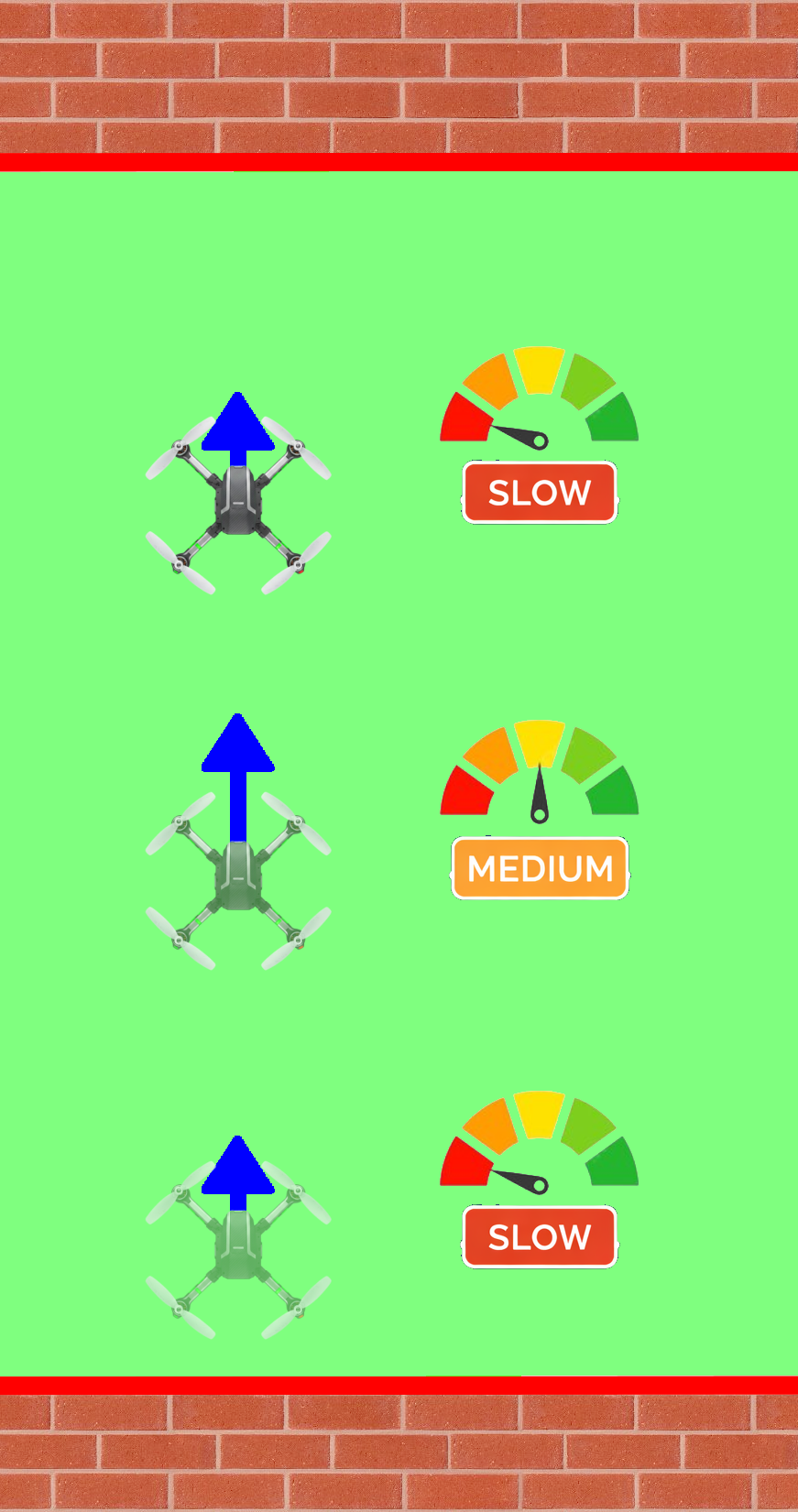}
  \end{subfigure}\hfill
  \begin{subfigure}[t]{0.31\columnwidth}
    \centering
    \includegraphics[width=0.95\columnwidth]{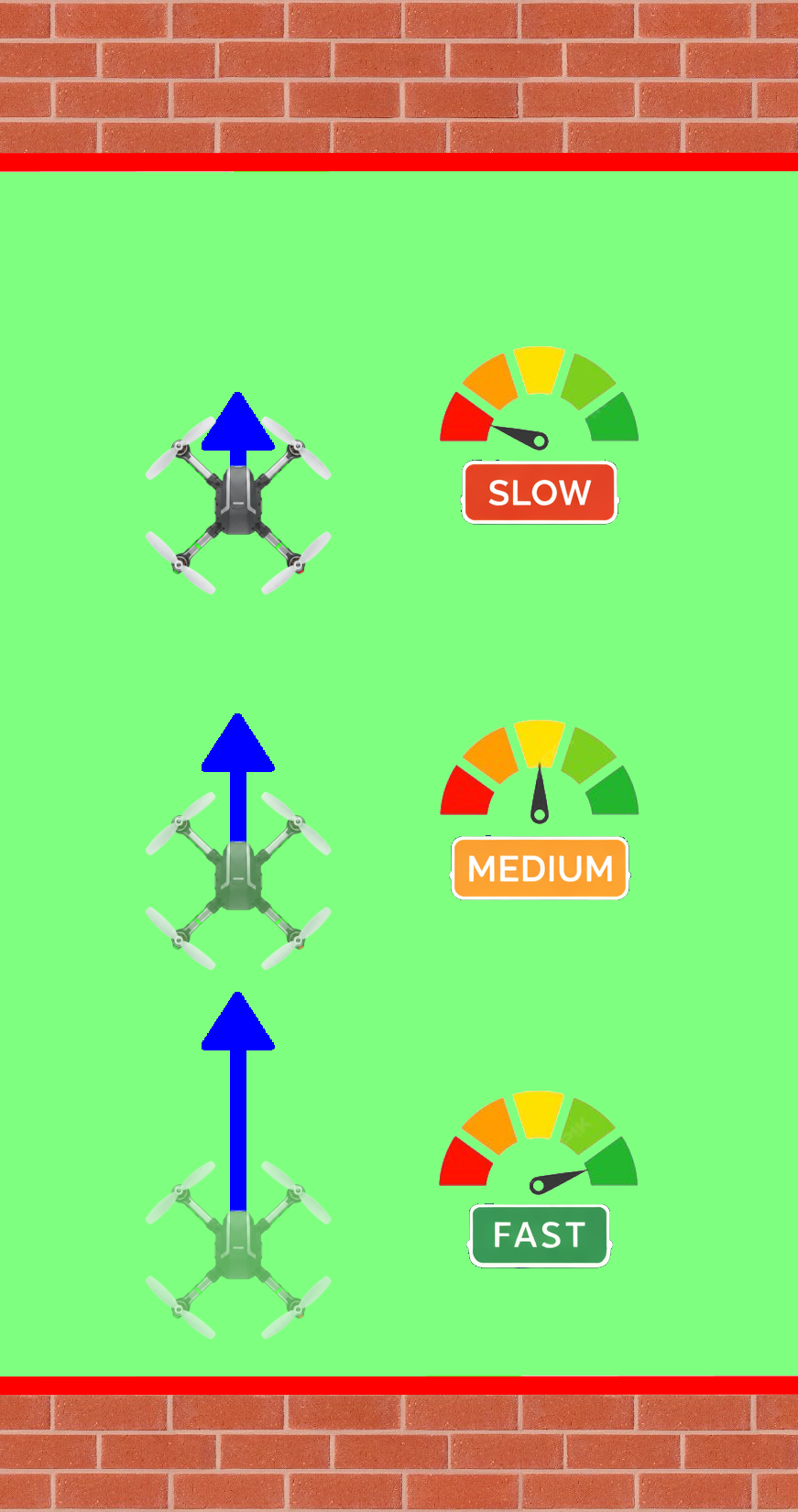}
  \end{subfigure}
  \begin{subfigure}[t]{0.31\columnwidth}
  \vspace{0pt}
    \centering
    \includegraphics[width=0.95\columnwidth]{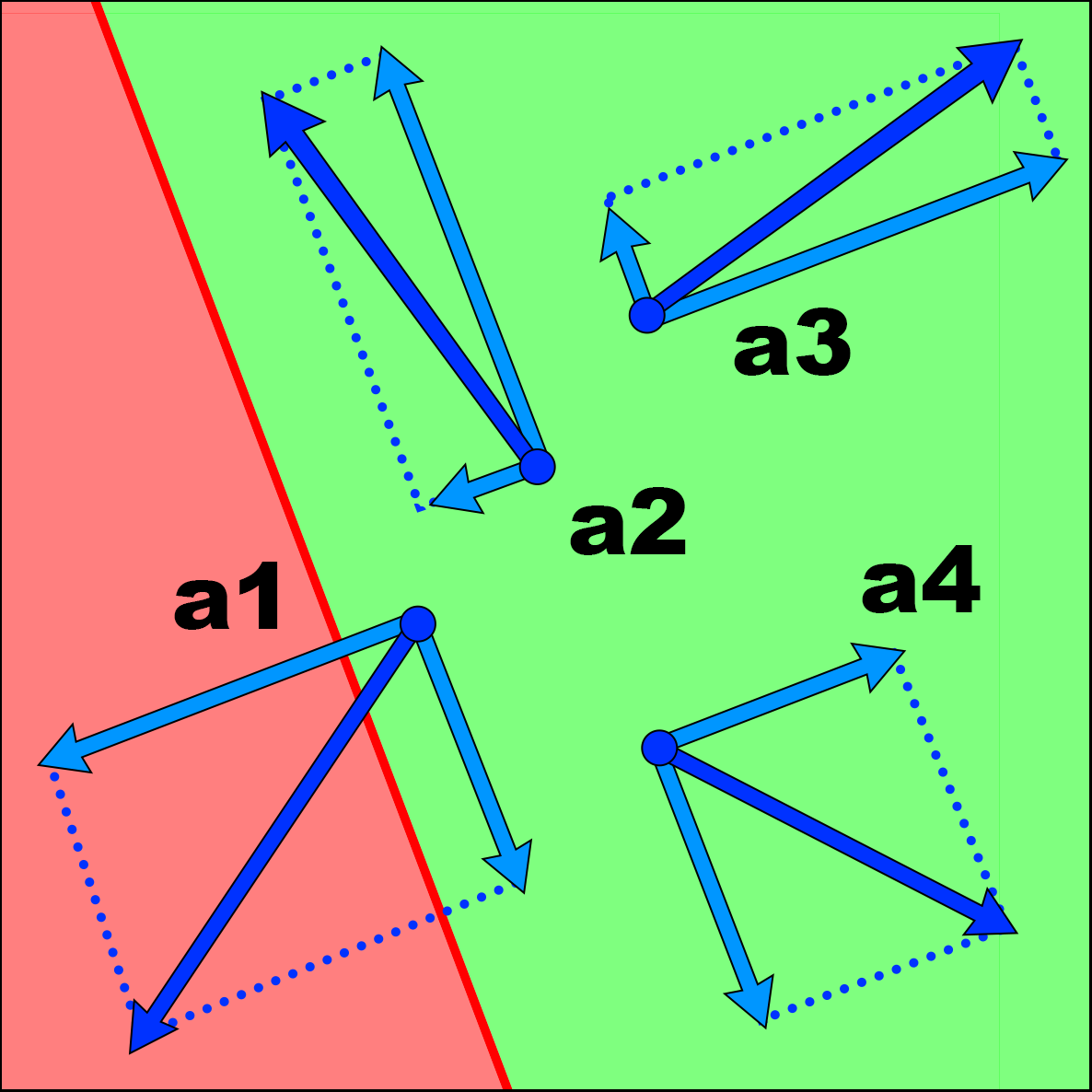}
    \caption{Unconstrained}
  \end{subfigure}\hfill
  \begin{subfigure}[t]{0.31\columnwidth}
  \vspace{0pt}
    \centering
    \includegraphics[width=0.95\columnwidth]{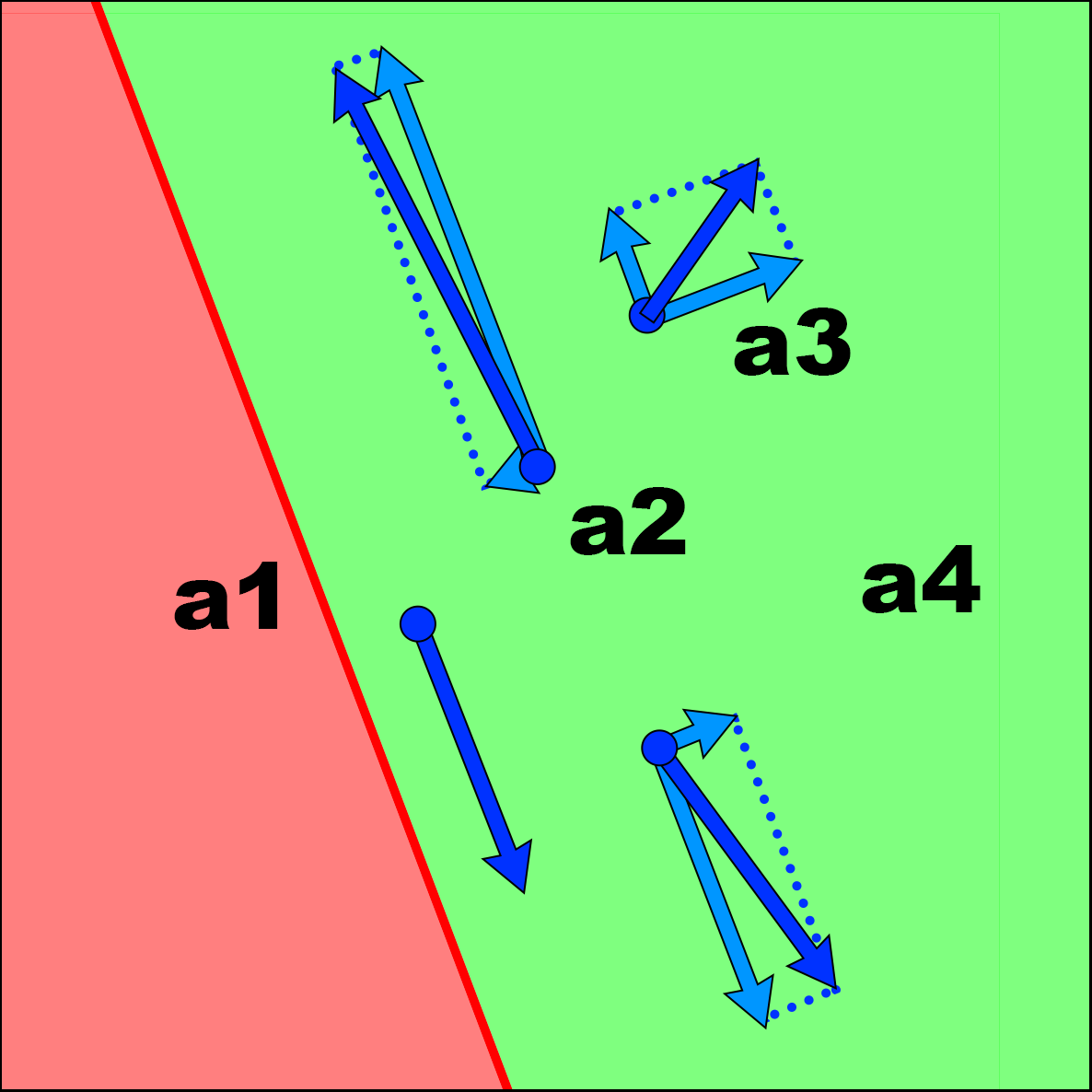}
    \caption{ATACOM~\cite{liu2025safe}}
  \end{subfigure}\hfill
  \begin{subfigure}[t]{0.31\columnwidth}
  \vspace{0pt}
    \centering
    \includegraphics[width=0.95\columnwidth]{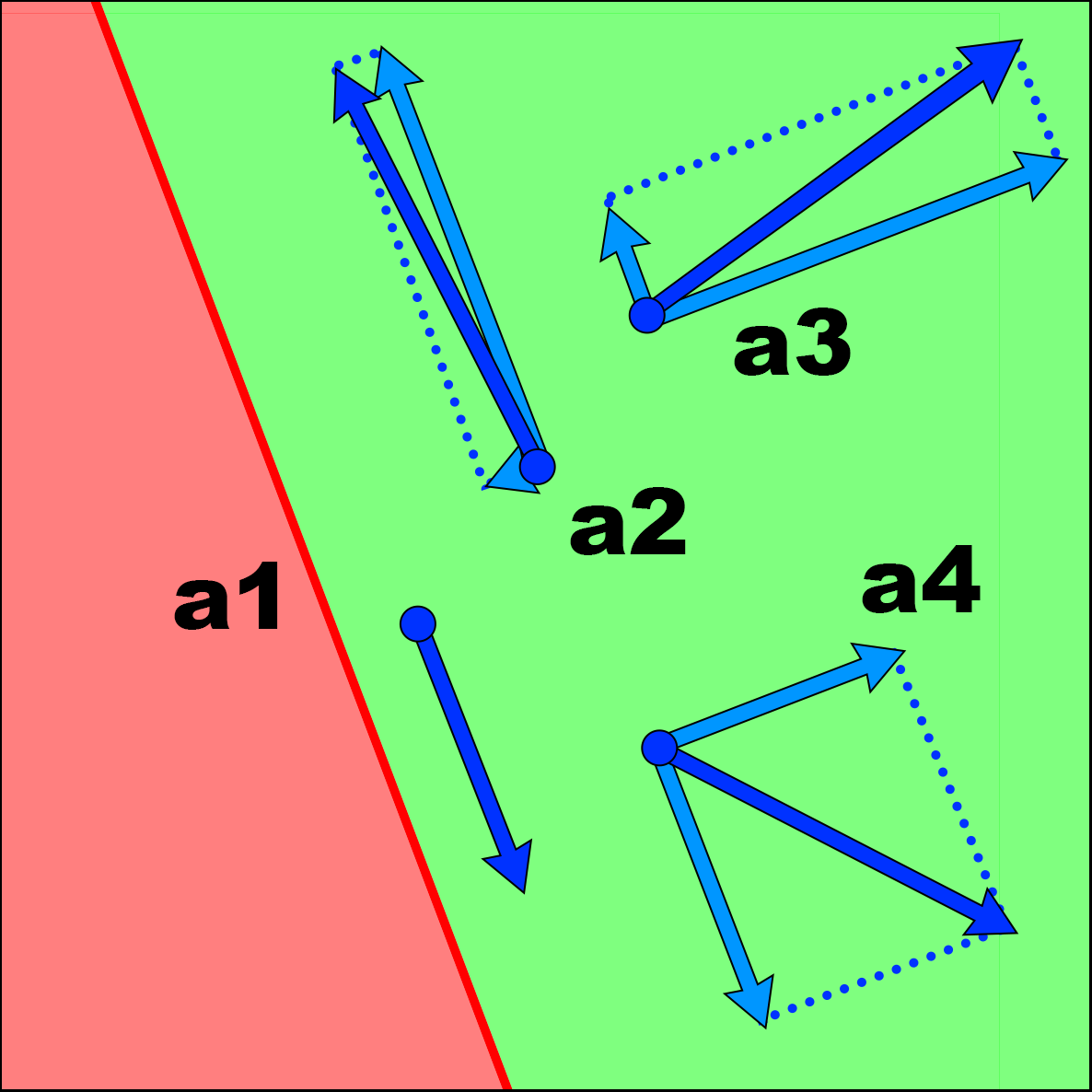}
    \caption{Directional (ours)}
  \end{subfigure}
    \caption{
    Illustration of four actions near a constraint (red line) under three handling strategies:
    (a) raw agent actions without modification;
    (b) scaling of the orthogonal component for all actions while preserving tangential components;
    (c) selective scaling of only actions directed toward the constraint (a1, a2), leaving safe actions (a3, a4) unchanged.
    }

  \label{fig:directional}
  \vspace{-2em}
\end{figure}

To tackle this problem, both data-driven \gls{saferl}~\cite{gu2024review} methods based on a constraint budget, and model-based \gls{safeexp}~\cite{garcia2012safe,berkenkamp2017safe} 
approaches based on state constraints have been developed. 
While each of these methods has different strengths, weaknesses, and application domains, there is a common issue to both settings: learning under safety constraints is considerably more challenging than the standard unconstrained \gls{rl} problem. 
Indeed, trying to satisfy safety requirements while optimizing a policy may be challenging, either because the policy objective could push the system towards constraint violations---e.g., when extreme motions are more effective than slower and safer ones---or because the safety constraints restrict exploration too much.
Learning is even more problematic when constraint functions are unknown and must be estimated from data, since the additional function-approximation process may further restrict exploration.

In this paper, we address the issue of efficient exploration under safety constraints.
Our work is based on one of the state-of-the-art approaches for learning under safety constraints, the \gls{atacom}~\cite{liu2022robot} framework, which has been recently~\cite{liu2025safe} shown to be capable of performing real-world finetuning of RL policies on complex, dynamic, contact-rich manipulation tasks such as the Robot Air Hockey.
This approach has originally been developed in the context of \gls{safeexp}, being closely related to \gls{cbf}~\cite{ames2019control,liu2025safe} and later extended in the context of \gls{saferl}~\cite{guenster2024handling} to include distributional critics~\cite{yang2021wcsac,yang2023safety} and \gls{fvf}~\cite{yang2023feasible}.

While the \gls{atacom} approach is effective in maintaining safety and has a low computational overhead, only requiring simple matrix operations and no optimization, the safety layer hampers exploration, making learning slower and more inefficient, particularly during the first training epochs.
To address this issue, we introduce the concept of \textit{Directional Constraints}, which morphs the action through the safety layer only when approaching the constraints, while leaving it unchanged when moving away from them, thereby promoting safe exploration.
A comparison between the original~\cite{liu2025safe} and our Directional Constraints approach is illustrated in Figure~\ref{fig:directional}. 
Both (b) and (c) ensure constraint satisfaction, unlike case (a), where \gls{atacom} is not applied. Additionally, Directional Constraints (c) selectively scale down only the actions that move towards the constraint. In contrast, the original \gls{atacom}  (b) also modifies outbound actions, forcing the agent to stay closer to the boundary even when unnecessary.
The contributions of this work are summarized as follows: 
\begin{itemize}
     \item We introduce \gls{atacomdc}, a novel approach to generate safe actions. This is a simple, yet very effective modification of the \gls{atacom} safety layer, which allows significantly faster learning and, in some cases, achieve better final performances in multiple challenging robotic control benchmark tasks.
    \item Furthermore, we systematically analyze the sensitivity of the parameters of \gls{atacom} and \gls{atacomdc}, highlighting the performance trade-off between the policy's safety and performance.
\end{itemize}

\section{Related Work}
Many approaches have been developed to deal with real-world safety-critical settings in learning-based control. 
Machine learning-based \gls{saferl} methods are built on the \gls{cmdp} framework~\cite{altman1998constrained}. The key idea of \gls{saferl} is to impose safety constraints, represented as a threshold on a cumulative constraint cost during the learning process. These approaches are generally based either on the idea of safety filters~\cite{dalal2018safe,nguyen2024gameplay} or on Constrained optimization~\cite{achiam2017constrained,liu2020ipo,ding2021provably}, among which the family of Lagrangian-based methods~\cite{ray2019benchmarking,ha2021learning,yang2021wcsac,yang2023safety} is the most represented.

In the fields of robotics and control, major efforts have been made to advance the \gls{safeexp} area,
due to its closeness to real-world applications.
To tackle the \gls{safeexp} problem, a variety of approaches have been developed based on the theory of \gls{cbf}~\cite{ames2019control,taylor2020learning,xiao2022high_order,tan2023your,yang2023model}, Lyapunov Stability~\cite{chow2018lyapunov,chow2019lyapunov}, Reachability Analysis~\cite{akametalu2014reachability,fisac2018general,shao2021reachability}, and Model Predictive Control~\cite{hewing2020learning,prajapat2024towards,prajapat2025safe}.
Furthermore, data-driven solutions~\cite{berkenkamp2017safe,wendl2026safe} tackle the problem by exploiting the regularity of the environment, such as Lipschitz-continuity and regularity of the unknown dynamics.

Machine learning and robot control approaches have both strengths and weaknesses. 
Machine learning-based~\gls{saferl} approaches are unable to impose safety at every timestep of the interaction with the environment, as safety is guaranteed (in probability) only at convergence.
Instead, control-based~\gls{safeexp} methods are usually not plug-and-play, since they rely on prior knowledge in the form of accurate dynamics models or hand-crafted functions.
Moreover, step-by-step safety guarantees may come at the cost of increased computations and overly conservative exploration.
In this work, we successfully tackle some of these limitations by directionally activating constraints, speeding up safe exploration.
\section{Preliminaries}
\label{sec:background}

\subsection{Constrained Markov Decision Process }

In \gls{saferl}, the environment is modeled as a \gls{cmdp}. A \gls{cmdp} is defined by the tuple $\langle \mathcal{S}, \mathcal{A}, \mathcal{R}, \mathcal{P}, \iota, \gamma, \mathcal{K} \rangle$, where $\langle \mathcal{S}, \mathcal{A}, \mathcal{R}, \mathcal{P}, \iota, \gamma \rangle$ corresponds to a standard MDP, and $\mathcal{K}$ is the set of constraint functions:
\begin{equation*}
    \mathcal{K} := \{ k_i : \mathcal{S} \rightarrow \mathbb{R} \mid
    i \in \{1, \dots, K\} \}.
\end{equation*}

In this paper, we will focus on  \gls{safeexp}, where we must ensure safety at every timestep of the agent-environment interaction.
To ensure safe exploration, the objective is to avoid constraint violations throughout the learning process. This is achieved by solving the following constrained optimization problem:

\begin{equation*}
\begin{aligned}
\pi^* &= \arg\max_{\pi} \, \mathbb{E}_{\tau \sim \pi}
\left[ \sum_{t=1}^{T} \gamma^t r(s_t, a_t) \right], \\
&\text{s.t. } k_i(s_t) \leq 0, \ \forall t, \forall i \in \{1, \dots, K\}.
\end{aligned}
\end{equation*}

\subsection{Acting on the Tangent Space of the Constraint Manifold}
The \gls{atacom} framework~\cite{liu2022robot,liu2023safe,liu2025safe} provides a method that achieves safe exploration by leveraging knowledge of the system's dynamics and constraints. 
At its core, ATACOM introduces the notion of \emph{constraint manifold}, transforming the original constrained optimization problem into an unconstrained one defined over this manifold.
To maintain safety, the action space is redefined as the tangent space of the constraint manifold. This ensures that any selected action satisfies the system constraints, as it lies within the safe set, i.e., the tangent space.

Based on the definition of the constraints, the safe region within the state space is the set $\mathcal{C} $,
\begin{equation*}
    \mathcal{C} := \left\{ \vs \in \mathcal{S} \subset \mathbb{R}^n \mid k(\vs) \leq 0 \right\}.
\end{equation*}
To build the constraint manifold, the state space is augmented with a vector of \emph{slack variables} $\vmu \in [0, +\infty)^K $. This leads to the augmented constraint function:
\begin{equation*}
    \vc(\vs, \vmu) := \mathbf{k}(\vs) + \vmu.
\end{equation*}
The corresponding constraint manifold is then defined as the set of all state-slack pairs that satisfy the equality condition:
\begin{equation*}
    \mathcal{M} := \left\{ (\vs, \vmu) \in \mathcal{D} \mid \vc(\vs, \vmu) = \mathbf{0} \right\}.
\end{equation*}

ATACOM constructs a safety controller that ensures that the system evolves within the tangent space of the constraint manifold. To achieve this, it assumes that the system can be modeled as a control-affine dynamical system
\begin{equation}
    \dot{\vs} = \vf(\vs) + \vG(\vs) \vu_s
    \label{eq:affine_system}
\end{equation}
and that the constraints $k(\vs)$, along with their Jacobians $\vJ_k(\vs)$, are defined analytically. \gls{atacom} also defines per-element dynamics for the slack variables, using a class $\mathcal{K}$ function $\alpha_i(\cdot)$ which is locally Lipschitz continuous. In this paper, we use exponential slack dynamics
\begin{align*}
     \mu_i = \alpha_i(\mu_i)u_{\mu,i}, &  & \alpha_i(\mu_i) = \exp(\beta\mu_i) - 1.
\end{align*} 
The $\beta$ parameter is used by \gls{atacom} to perform a trade-off between safety and performance, with lower values of $\beta$ representing constraints with wider margin.

The controller is designed to adjust the augmented state along directions that remain tangent to the constraint manifold. In other words, the velocity vector $\begin{bmatrix} \dot{\vs} & \dot{\vmu} \end{bmatrix}^\top $must belong to the tangent space $\mathcal{T}_{(\vs,\vmu)} \mathcal{M} $.
Under this condition, the safe control input is defined as:
\begin{equation}
    \label{eq:atacom_controller}
    \begin{bmatrix}
    \vu_s \\
    \vu_\mu
    \end{bmatrix}
    = -\vJ_u^\dagger \boldsymbol{\psi} - \lambda \vJ_u^\dagger \vc + \vB_u \vu,
\end{equation}
where $\bm{\psi} = \vJ_k(\vs) \vf(\vs )$ is the constraint drift that describes how the constraints evolve without control action; $\vJ_u^\dagger$ denotes the pseudoinverse of 
$\vJ_u =\begin{bmatrix} J_k(\vs) G(\vs) & A(\bm{\mu}) \end{bmatrix}$.
Here, $A(\bm{\mu}) : \mathbb{R}^K \rightarrow \mathbb{R}^{K \times K}$ is a diagonal matrix with entries $A_{ii} = \alpha_i(\mu_i)$, while $\vB_u(\vs,\boldsymbol{\mu})$ is the tangent space basis in matrix form, such that the following equation holds 
\begin{equation*}
    \vJ_u(\vs,\boldsymbol{\mu}) \vB_u(\vs,\boldsymbol{\mu}) = \mathbf{0}.
\end{equation*}
The first term $\bm{J}_u^\dagger \bm{\psi}$ of the \gls{atacom} controller represents the drift compensation term, which counteracts the natural system drift. The second term $\lambda \bm{J}_u^\dagger \vc$ corresponds to the contraction term, driving the system state back toward the constraint manifold when the constraint is violated. The final component $\bm{B}_u \vu$ is the tangential term, generating a vector field that lies in the tangent space of the constraint manifold. Overall, the controller maps the task-specific control input $\vu$ into a safe action $\vu_s$.

\subsection{Distributional ATACOM}
\gls{datacom} is an extension~\cite{guenster2024handling} of the \gls{atacom} framework, removing the necessity of the analytical form of the constraints function. Instead, \gls{datacom} introduces the concept of \gls{fvf}, which evaluates the expected constraint violation under the policy.
To incorporate the uncertainty arising from various sources of stochasticity in the environment, the \gls{fvf} is learned using distributional \gls{rl}. The algorithm exploits the uncertainty estimate to improve safety through the concept of \gls{cvar}. Furthermore, the algorithm introduces a dynamic threshold $\delta$ to counteract constraint function fitting errors, ensuring that the system reaches a predefined value of safety while allowing for meaningful exploration in the initial phase of training.

\label{sec:methods}
While \gls{atacom} performs effectively on a wide range of tasks, it can produce suboptimal policies in scenarios where multiple constraints significantly restrict the action space. In such settings, the agent is forced to operate conservatively, which can negatively affect task execution and learning efficiency.
By design, \gls{atacom} suppresses the action component along the orthogonal direction of the constraint, thereby limiting motion toward constraint boundaries while allowing the \gls{rl} agent to freely explore the tangent space. While this mechanism effectively prevents constraint violations, it also unintentionally suppresses actions that would move the agent away from the constraint boundary.
As a consequence, once the agent approaches a constraint, escaping toward safer regions of the state space becomes difficult. This limitation arises from the symmetric nature of the action space morphing near constraint boundaries, which penalizes both approaching and escaping motions. 

\subsection{Key idea}
To address the issue of symmetric morphing, we implement the \emph{Directional Constraints} mechanism. 
The key idea is to scale down only the actions that are moving the robot towards the constraints. 
To decide which actions to scale, we evaluate the constraint derivatives, which describe how the constraint values evolve based on the current action. If the derivative is positive, the constraint value is increasing; otherwise, it is decreasing. To evaluate the effect of an action on the $i$-th constraint, we  consider the constraint derivative $\dot{c}_i$, as follows: \vspace{-.25em}
\begin{equation*}
    \dot{c}^i(\vs) = \psi^i(\vs) + \vJ^i_k(\vs) \vG(\vs) \vu_s,
\end{equation*}
where $\vJ^i_k(s)$ is the Jacobian of the $i$-th constraint, $\vG(s)$ is the function that models the system, and $\vu_s$ is the sampled action.
Once the constraints derivatives are computed, we can select the actions to scale based on the derivative's sign, only keeping the one with positive sign.
Thus, Directional Constraints apply the action morphing only if the  action sampled from the \gls{rl} policy would move the robot closer to the constraint, otherwise leaving  the action unaffected.

\begin{figure}[t]
  \centering
  \begin{subfigure}[t]{0.25\columnwidth}
    \centering
    \includegraphics[width=\columnwidth]{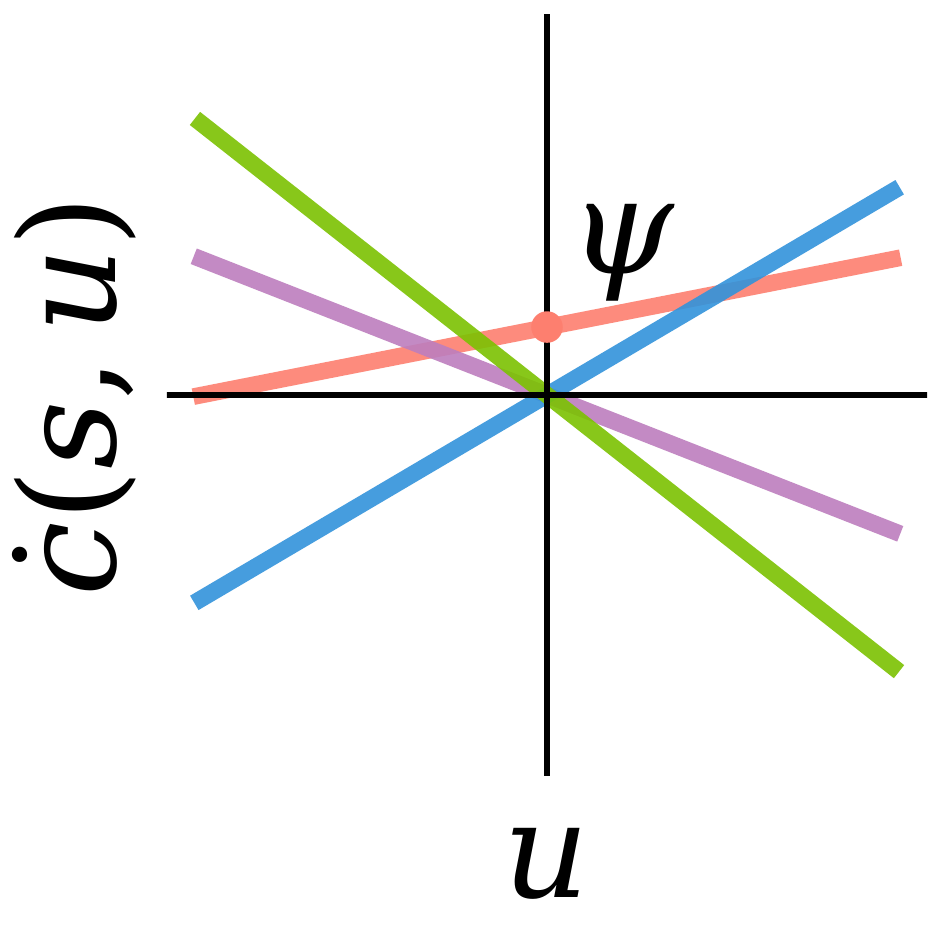}
    \caption{}
    \label{fig:constraint_derivative}
  \end{subfigure}\hfill
  \begin{subfigure}[t]{0.1\columnwidth}
    \centering
    \includegraphics[width=\columnwidth]{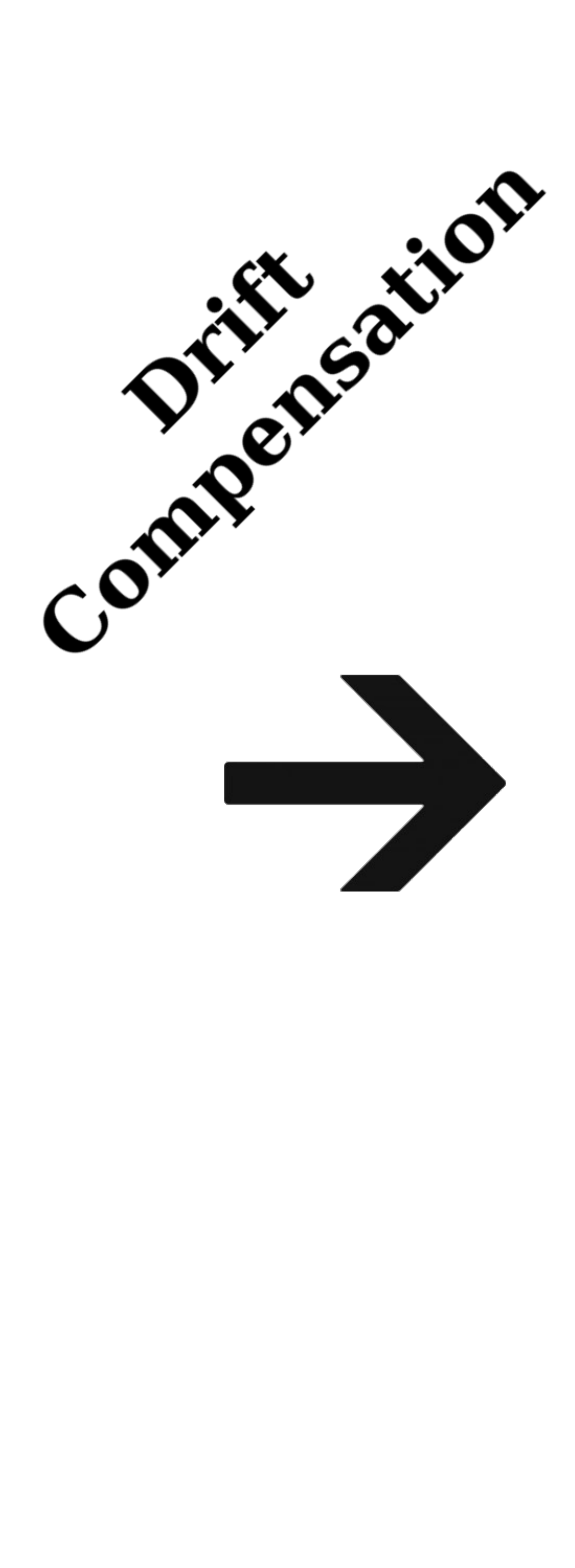}
  \end{subfigure}\hfill
  \begin{subfigure}[t]{0.25\columnwidth}
    \centering
    \includegraphics[width=\columnwidth]{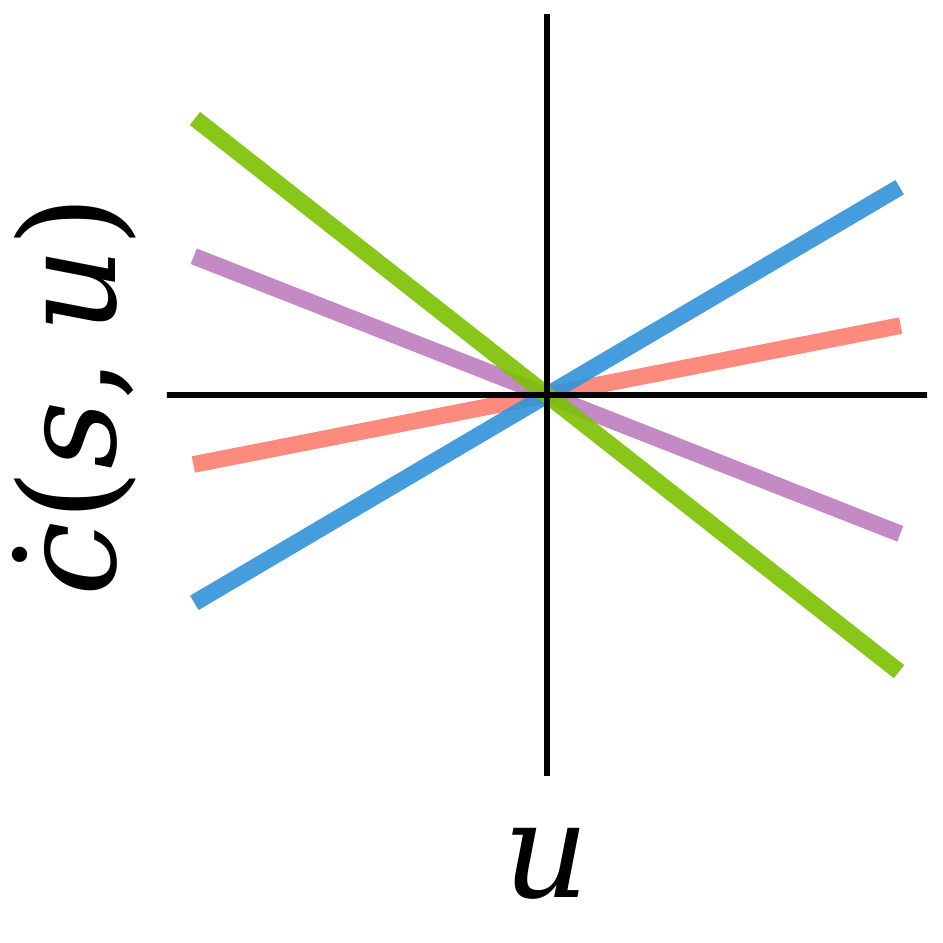}
    \caption{}
    \label{fig:constraint_derivative_drift_compensation}
  \end{subfigure}\hfill
  \begin{subfigure}[t]{0.1\columnwidth}
    \centering
    \includegraphics[width=\columnwidth]{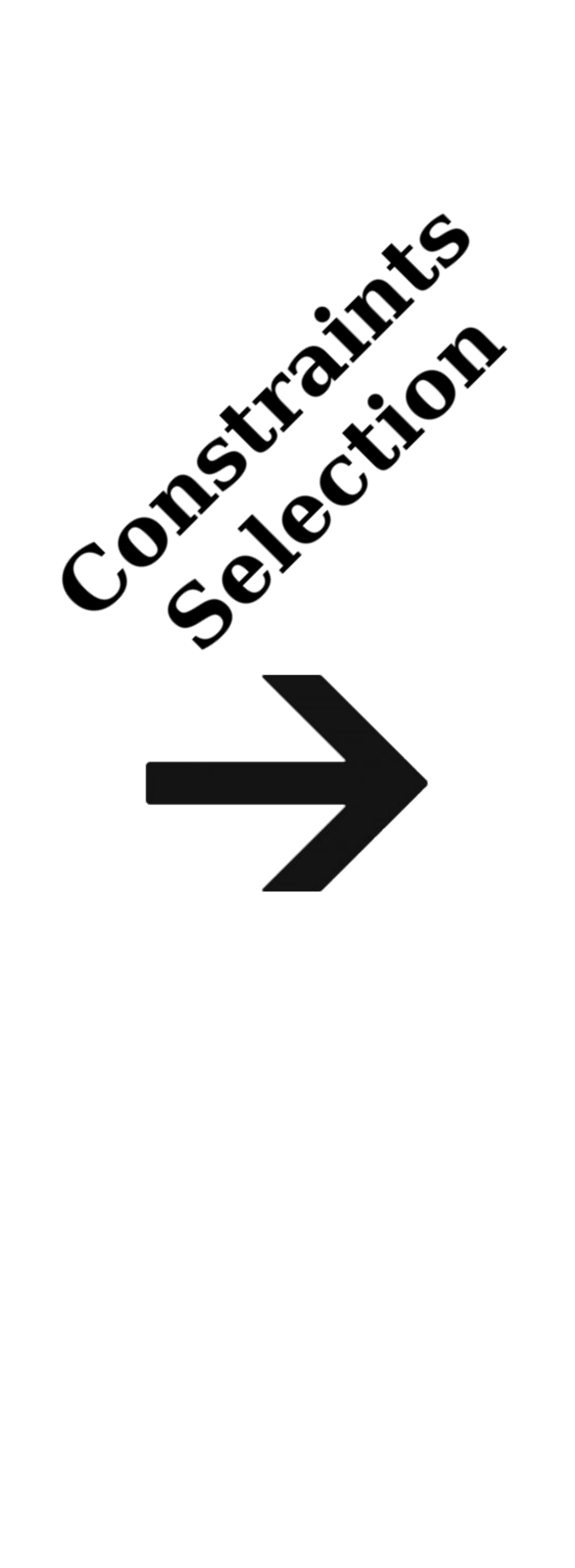}
  \end{subfigure}\hfill
  \begin{subfigure}[t]{0.25\columnwidth}
    \centering
    \includegraphics[width=\columnwidth]{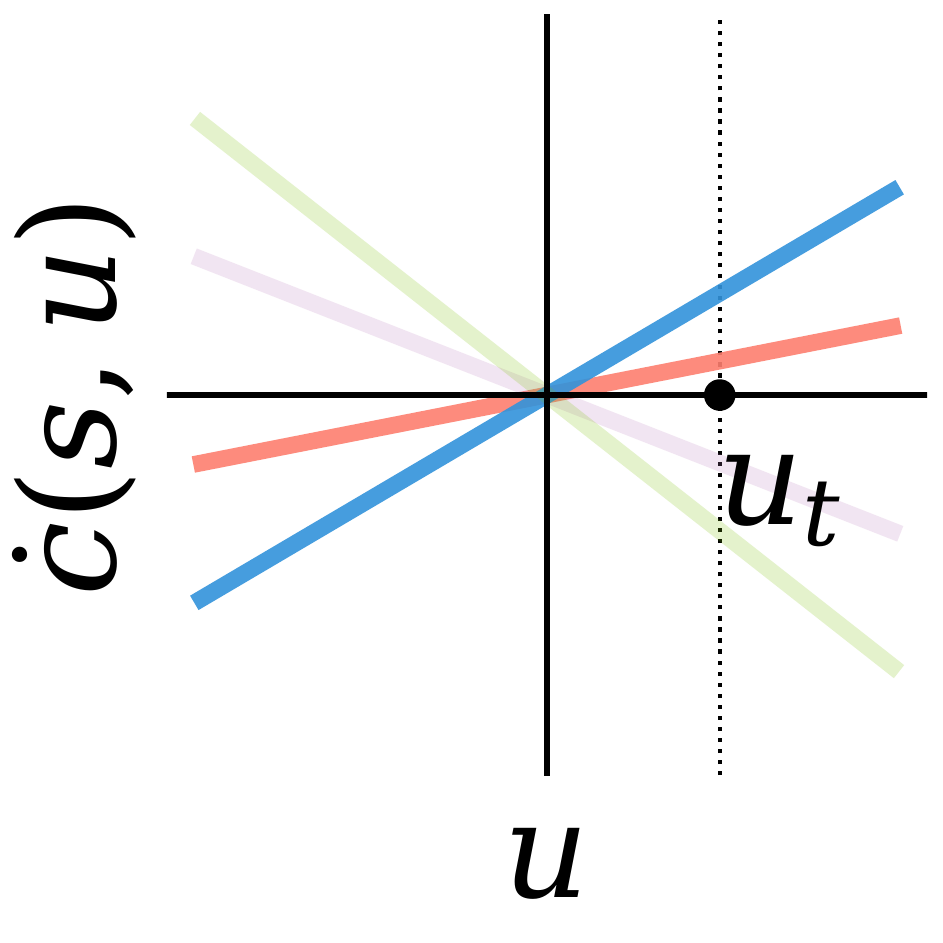}
    \caption{}
    \label{fig:constraint_derivative_action}
  \end{subfigure}
  \caption{
  A representation of the key idea of Directional Constraints with a one-dimensional action. Plots show the constraint derivatives $\dot{c}(\vs,\vu)$ for multiple constraints as a function of the sampled action. The system drift affects the constraint values for states close to the boundary, increasing the derivative even with zero control action. This effect is illustrated by the orange line, which intersects the y-axis at a nonzero value. After drift compensation, we sample the (residual) action $u_t$ and select constraints based on the sign of their derivative, disabling constraints with $\dot{c}(\vs,\vu) < 0$.}
  \vspace{-1em}
\end{figure}

\subsection{Multiple constraints}
Although the idea described above is intuitive for single constraints, extending it to multiple constraints is not straightforward. 
In the following, we exploit the geometry of the problem to make the extension viable for an arbitrary number of constraints.

First of all, we notice that all constraint derivatives are linear w.r.t. $u_s$. This means that each constraint derivative is a hyperplane in the space of actions, i.e., a line in the scalar-action setting, as shown in Figure~\ref{fig:constraint_derivative} as an illustrative example. 
As we are assuming a nonlinear affine system (\cref{eq:affine_system}), most hyperplanes (lines) will pass through the origin. 
In the \gls{atacom} setting, we assume that this linear system is solvable, as prescribed in Assumption 4 from~\cite{liu2025safe}
\begin{equation*}
    \vpsi(\vs) + \vJ_u(\vs,\boldsymbol{\mu}) 
    \begin{bmatrix} \vu_s \\ \vu_\mu \end{bmatrix} = \mathbf{0}.
\end{equation*}
This means that we can always compensate for the constraint drift. Notice that, while this assumption may seem restrictive, we are indeed restricting the drift compensation on the constraint manifold, which does not require complete compensation for the drift of the system. This means that the system drift is compensated only when reaching constraint boundaries; otherwise, the drift can be redirected into an increase of the slack variables.
For many practical scenarios, this assumption boils down to not having more active constraints, i.e., those with zero slack variable, than action degrees of freedom.
Therefore, in our setting, we can always compensate first for the drift of the system, and only learn the residual part of the controller, representing the ``tangent'' part of the action. 

With the drift compensated, all hyperplanes of the residual action will pass through the center of the plane, as illustrated in Figure~\ref{fig:constraint_derivative_drift_compensation}. We further assume that the null-space basis $\vB_u$ does not invert the sign of the actions of the actual system, which holds in most scenarios when using the smooth basis algorithm introduced in~\cite{liu2025safe}. Notice that this assumption only prescribes us to avoid the use of an inverting basis, i.e., a basis that changes the meaning of the action drastically. Given that the null-space basis can be chosen arbitrarily among all possible bases, this is not a strict assumption, as we can always use a basis that preserves the action's meaning. A possible problematic case is when there are equality constraints, and the Jacobian rank degenerates, e.g., when the constraint manifold has sphere-like topologies. In this scenario, the selection of a proper base is impossible due to the hairy-ball theorem~\cite{eisenberg1979proof}. However, these are edge cases, as most robotics systems either do not have such complex constraints or can be designed to circumvent the issue.

With these two assumptions, we see that the sampled ``tangential'' action prescribes a quadrant where our morphed action will also lie. Here, only the constraints active in this quadrant are relevant for the \gls{atacom} morphing.
Thus, these active constraints are the only constraints that we need to consider to compute the morphing, effectively turning off the others.
Intuitively, if the action pushes away from the constraints, we can avoid considering them.
The complete \gls{atacomdc}  algorithm consists of the following steps: 
\begin{enumerate}
    \item Compensate for the drift;
    \item Sample a residual action; 
    \item Compute the active set of constraints by computing the constraint derivatives;
    \item Compute the morphing of the action using the \gls{atacom} controller only on the active set of constraints.    
\end{enumerate}
Notice that this approach retains the same safety guarantees as the original algorithm under the assumption of a perfect model and constraints, since it modifies the algorithm's behavior only when taking safe actions.

\subsection{Practical Implementation}
\begin{algorithm}[t]
    \caption{\gls{atacomdc}}
    \label{alg:atacom_dc}
    \small
    \begin{algorithmic}[1]
        \REQUIRE $\vs$, $\vu$ \hfill $\triangleright$ At each step
        \STATE Determine the slack variable \\
        \hspace{1.2em} $\boldsymbol{\mu} \leftarrow \max(-\vk(\vs), \text{tol})$
        \STATE Compute the Jacobians and the drift \\
        \hspace{1.2em} $\vJ_G \leftarrow \vJ_k(\vs)\vG(\vs)$ \\
        \hspace{1.2em} $\vJ_u(\vs,\boldsymbol{\mu}) \leftarrow 
        \begin{bmatrix}\vJ_G(\vs) & A(\boldsymbol{\mu})\end{bmatrix}$  \\
        \hspace{1.2em} $\vpsi(\vs) \leftarrow \vJ_k(\vs)\vf(\vs)$
        \STATE Compute the constraint value and derivative assuming drift compensation \\
        \hspace{1.2em} $\vc(\vs, \boldsymbol{\mu}) \leftarrow \vk(\vs) + \boldsymbol{\mu}$ \\
        \hspace{1.2em} $\dot{\vc}(\vs,\vu) \leftarrow \vJ_G(\vs) \vu$
        \STATE Disable constraints with negative $\dot{\vc}(\vs,\vu)$ \\
        \hspace{1.2em} $A(\boldsymbol{\mu})^{dir} \leftarrow A(\boldsymbol{\mu})_{[i | \dot{\vc} > 0]}$
        \STATE Compute the directional Jacobian\\
        \hspace{1.2em} $\vJ_u(\vs,\boldsymbol{\mu})^{dir} \leftarrow 
        \begin{bmatrix}\vJ_G(\vs) & A(\boldsymbol{\mu})^{dir}\end{bmatrix}$  \\
        \STATE Compute the tangent space basis \\
        \hspace{1.2em} $\vB_u \leftarrow \text{SmoothBasis}(\vJ_u^{dir})$
        \STATE Compute $\vu_s$ compensating the drift \hfill $\triangleright$ Eq.~\eqref{eq:atacom_controller}
        \STATE \textbf{Output:} $\vu_s$
    \end{algorithmic}
\end{algorithm}
The implementation of Directional Constraints follows Algorithm~\ref{alg:atacom_dc} and is conceptually illustrated in Figure~\ref{fig:constraint_derivative_action}. 
Given the sampled action at time step $t$, the derivative of each constraint is computed and evaluated to disable those constraints whose value would not increase if the action were applied.
Once the constraint derivatives $\dot{\vc}(\vs,\vu)$ are computed, the key step consists of removing the rows of the matrix $A(\boldsymbol{\mu})$ corresponding to the constraints with negative derivative, under the assumption that drift compensation is handled by the \gls{atacom} controller. 
The modified Jacobian $J_u^{\mathrm{dir}}(\vs,\boldsymbol{\mu})$ is then constructed accordingly, together with the corresponding smooth basis $\vB_u$. 

For computational efficiency, especially in configurations with parallel environments where the matrix $A(\boldsymbol{\mu})$ is computed in batch form, explicitly removing rows becomes impractical. Different environments may require removing a different number of rows, leading to batched matrices with inconsistent dimensions. An equivalent and more practical solution consists in setting the corresponding diagonal entries to the upper bound value, i.e.,
$\alpha_i(\mu_i) = \mu_\eta$ for all $i$ such that $\dot{c}_i(\vs,\vu) < 0$, 
where $\mu_\eta < \infty$ is a sufficiently large constant such that $\mu \leq \mu_\eta$.

\section{Experimental Evaluation}
\label{sec:results}
For the experimental evaluation of the \gls{atacomdc} approach, we  focus on three simulated tasks that are slightly modified versions of those proposed in~\cite{liu2025safe} and~\cite{guenster2024handling}.
Policy rollouts and qualitative comparisons on the considered tasks are available in the supplementary video.

\begin{figure*}[t]
    \centering
    \begin{subfigure}[t]{0.29\textwidth}
        \centering
        \includegraphics[width=\textwidth]{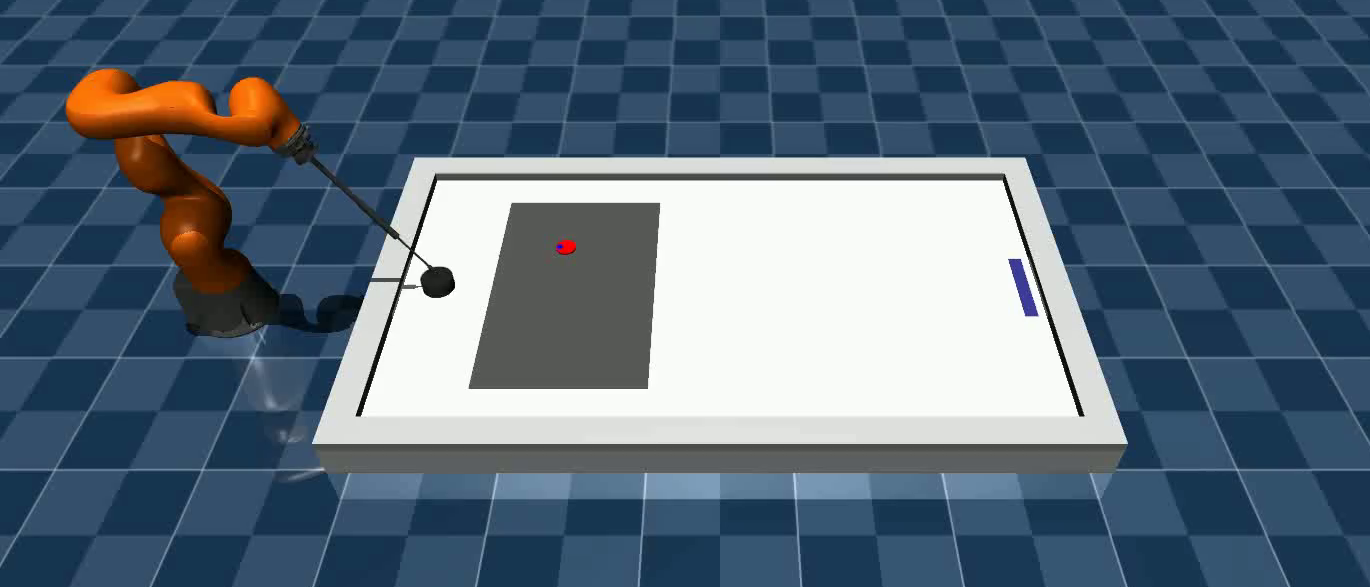}
        \caption{Kuka iiwa air hockey.}
    \end{subfigure}
    \begin{subfigure}[t]{0.29\textwidth}
        \centering
        \includegraphics[width=\textwidth]{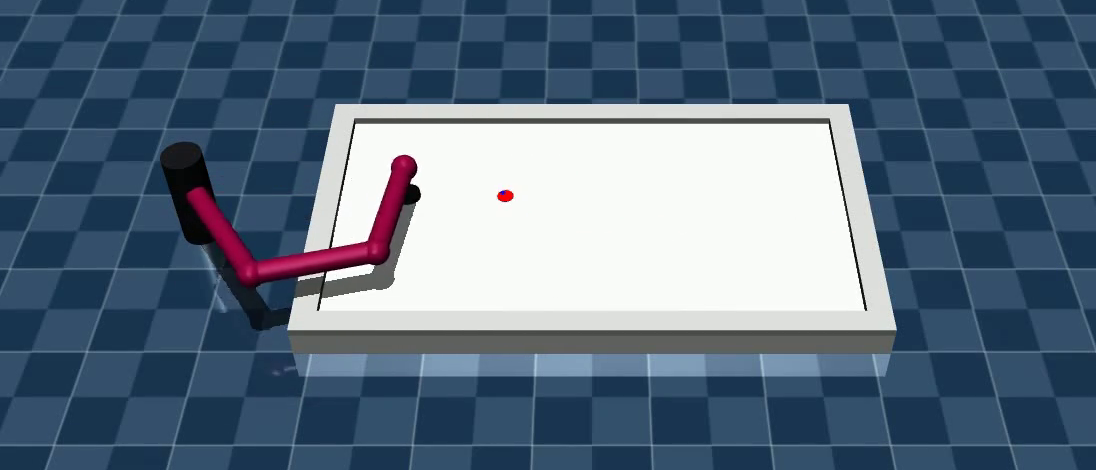}
        \caption{Planar air hockey.}
    \end{subfigure}
    \begin{subfigure}[t]{0.29\textwidth}
        \centering
        \includegraphics[width=\textwidth]{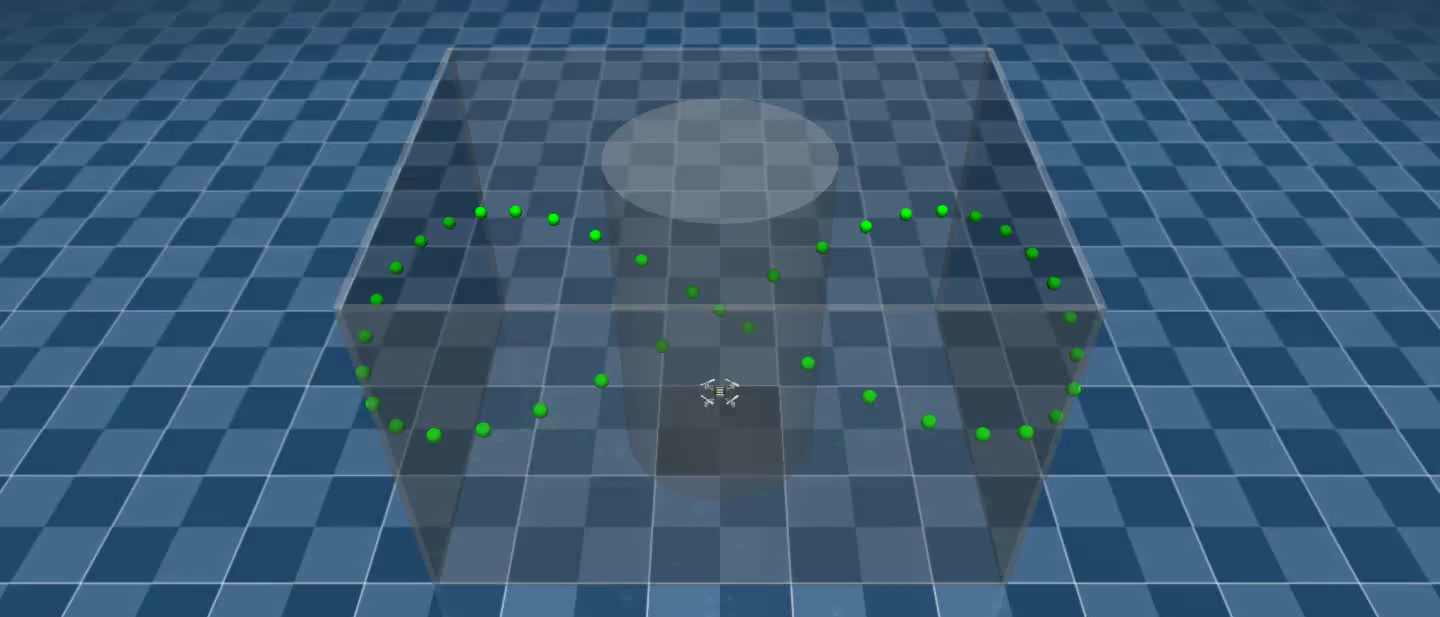}
        \caption{Quadrotor.}
    \end{subfigure}
    \caption{The task used in our experimental evaluation}
    \label{fig:tasks}
    \vspace{-1em}
\end{figure*}
\begin{figure*}[t]
    \centering
    \begin{subfigure}[t]{0.29\textwidth}
        \centering
        \includegraphics[width=\textwidth]{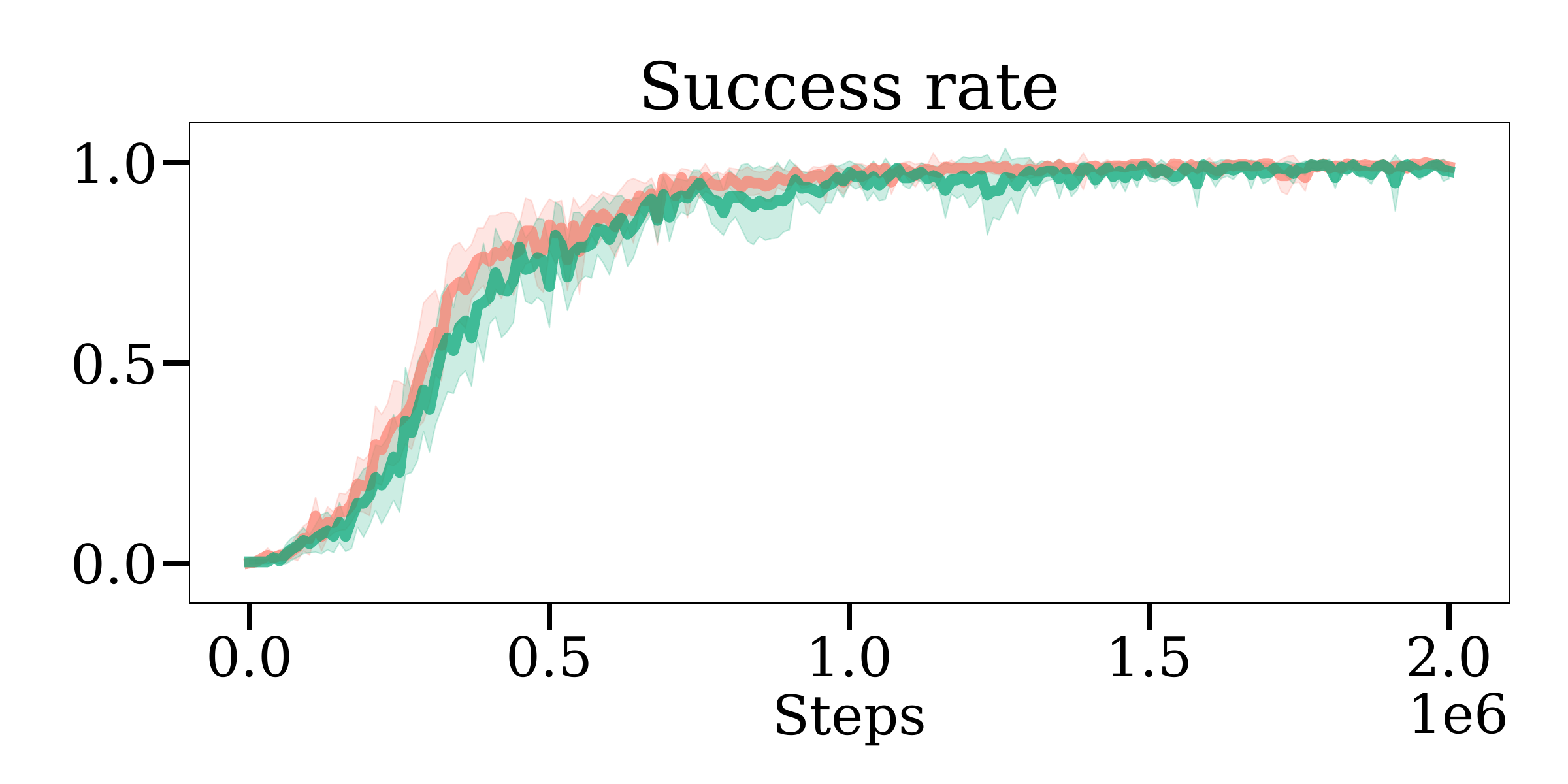}
    \end{subfigure}
    \begin{subfigure}[t]{0.29\textwidth}
        \centering
        \includegraphics[width=\textwidth]{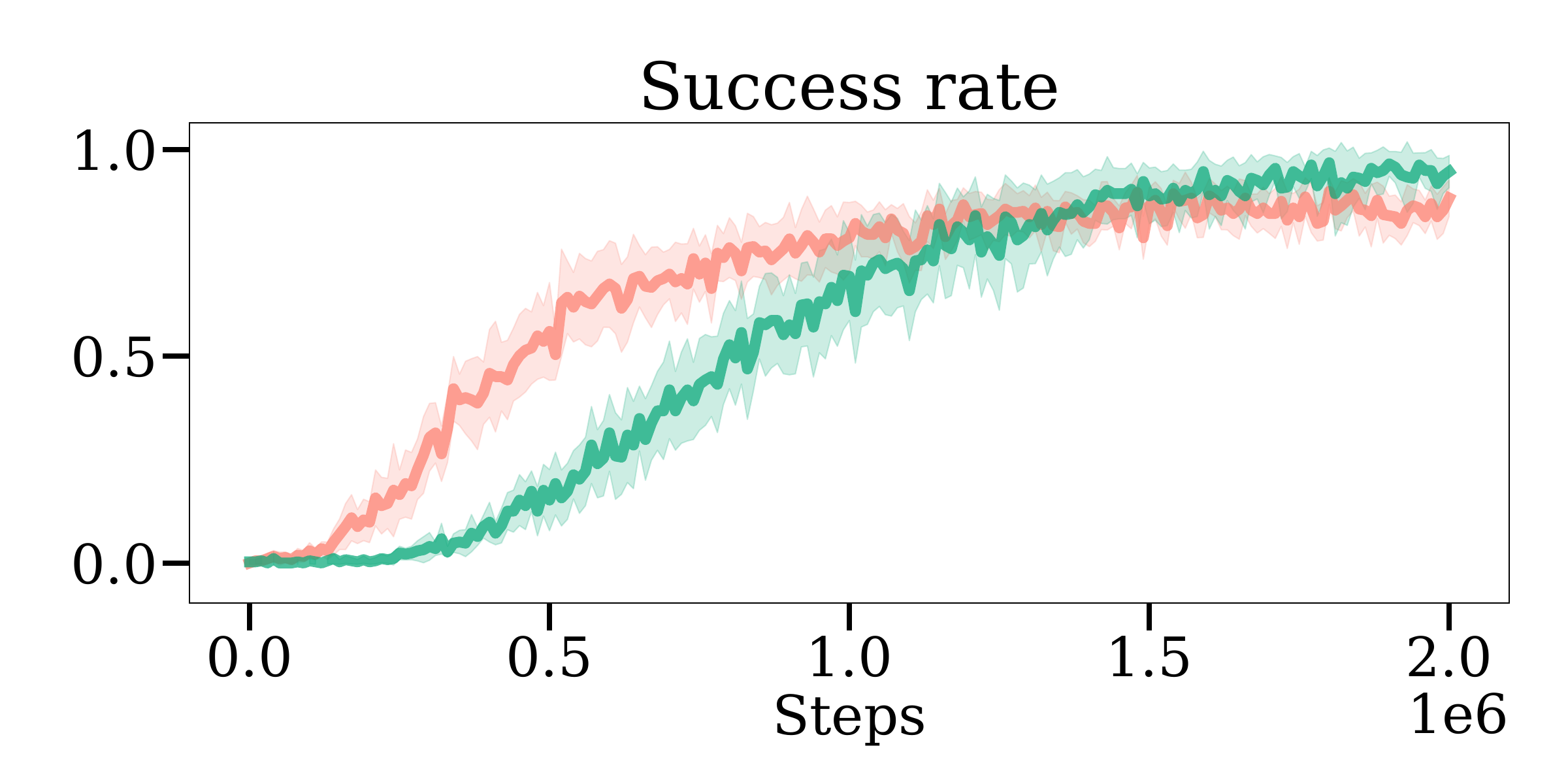}
    \end{subfigure}
    \begin{subfigure}[t]{0.29\textwidth}
        \centering
        \includegraphics[width=\textwidth]{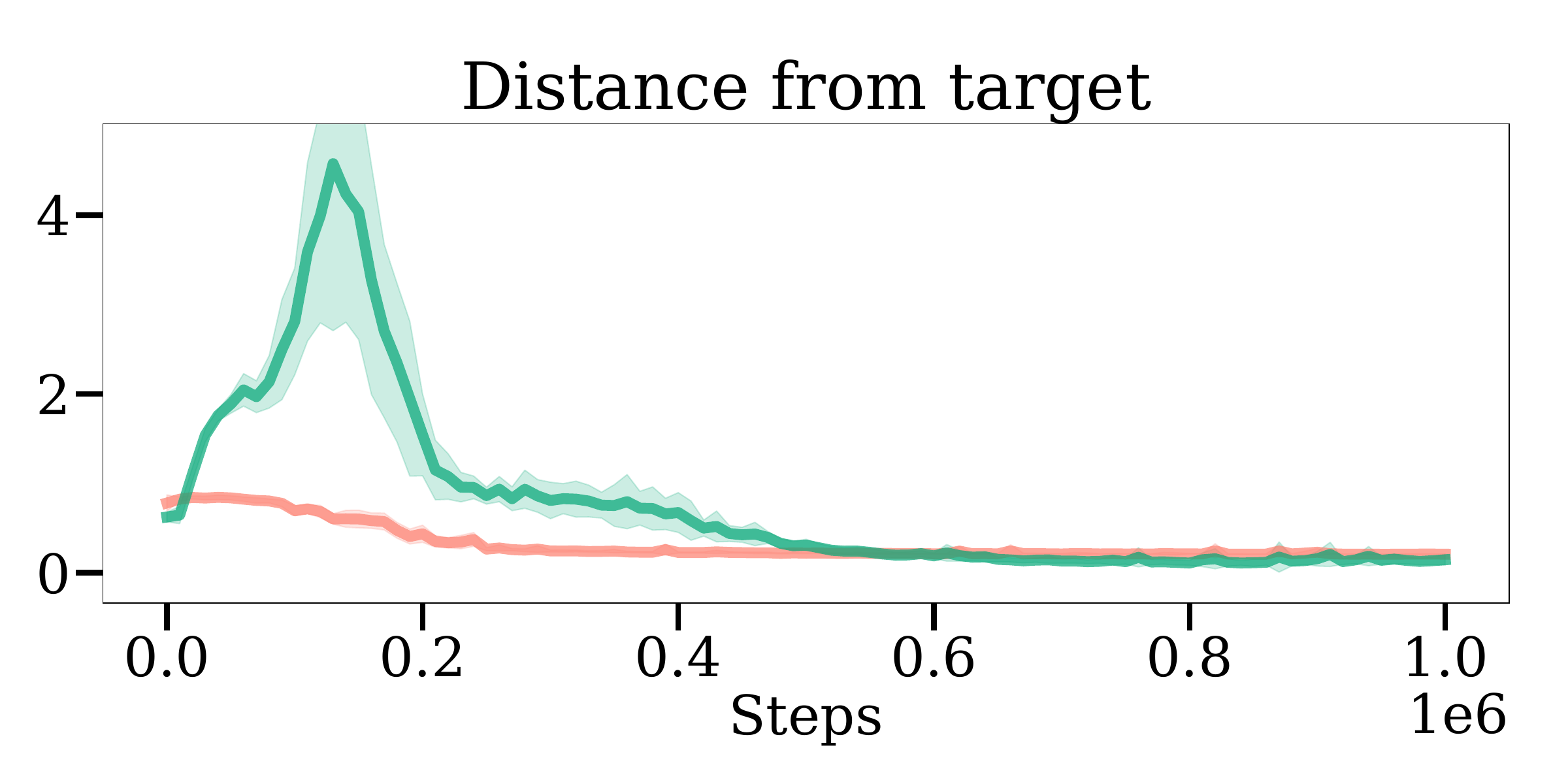}
    \end{subfigure}
    \begin{subfigure}[t]{0.29\textwidth}
        \centering
        \includegraphics[width=\textwidth]{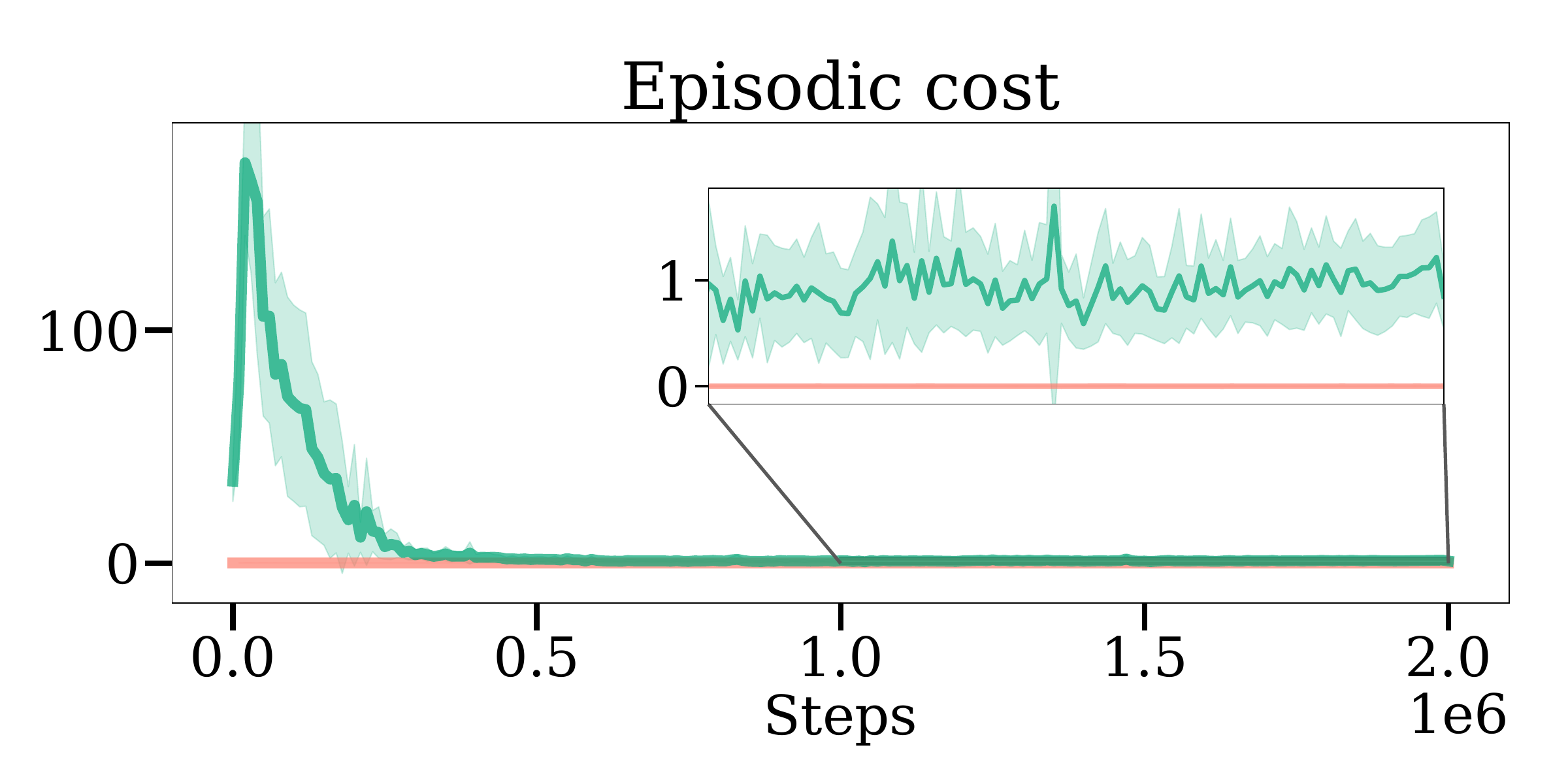}
        \caption{Planar air hockey (velocity)}
    \end{subfigure}
    \begin{subfigure}[t]{0.29\textwidth}
        \centering
        \includegraphics[width=\textwidth]{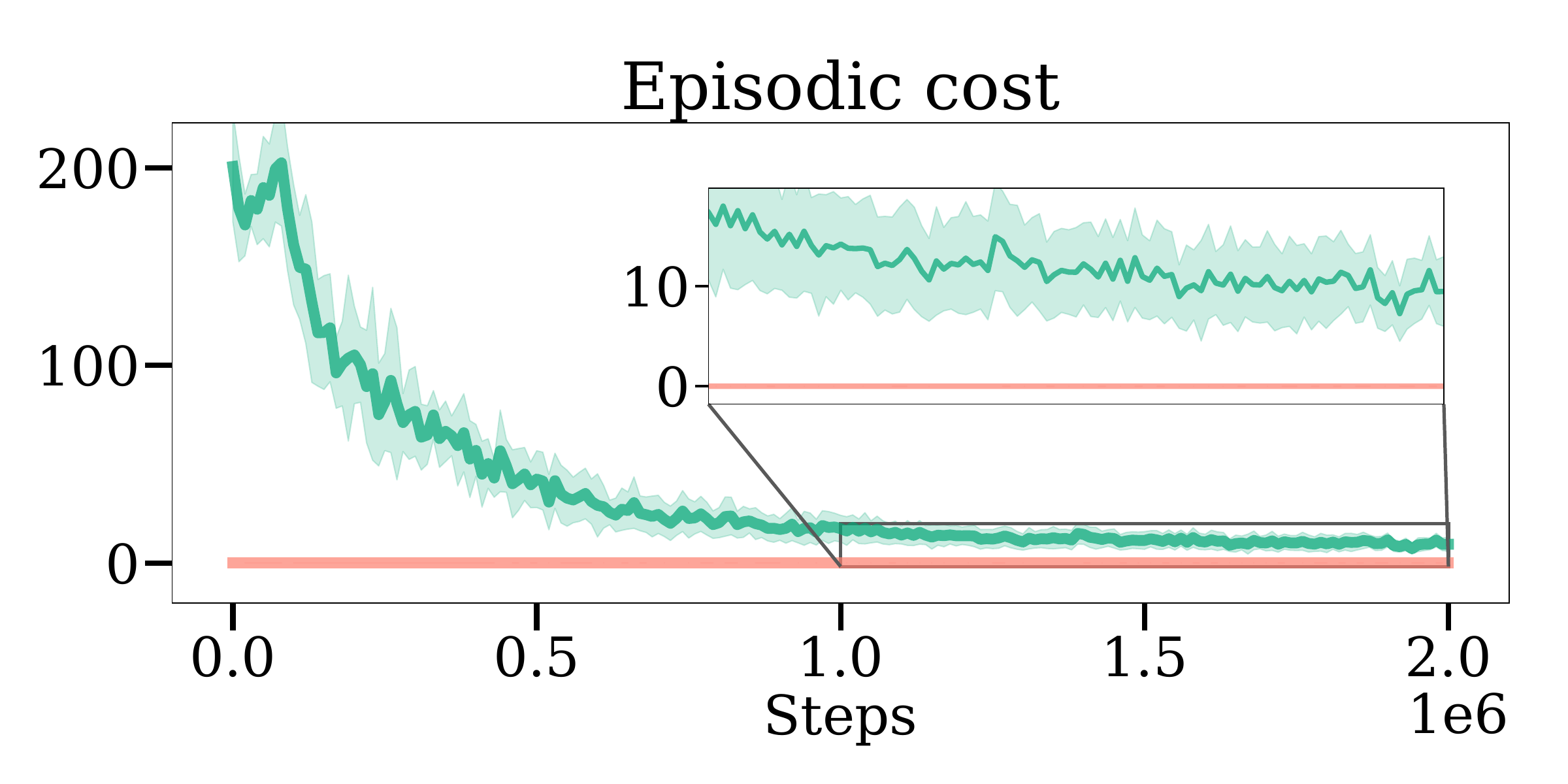}
        \caption{Planar air hockey (acceleration)}
    \end{subfigure}
    \begin{subfigure}[t]{0.29\textwidth}
        \centering
        \includegraphics[width=\textwidth]{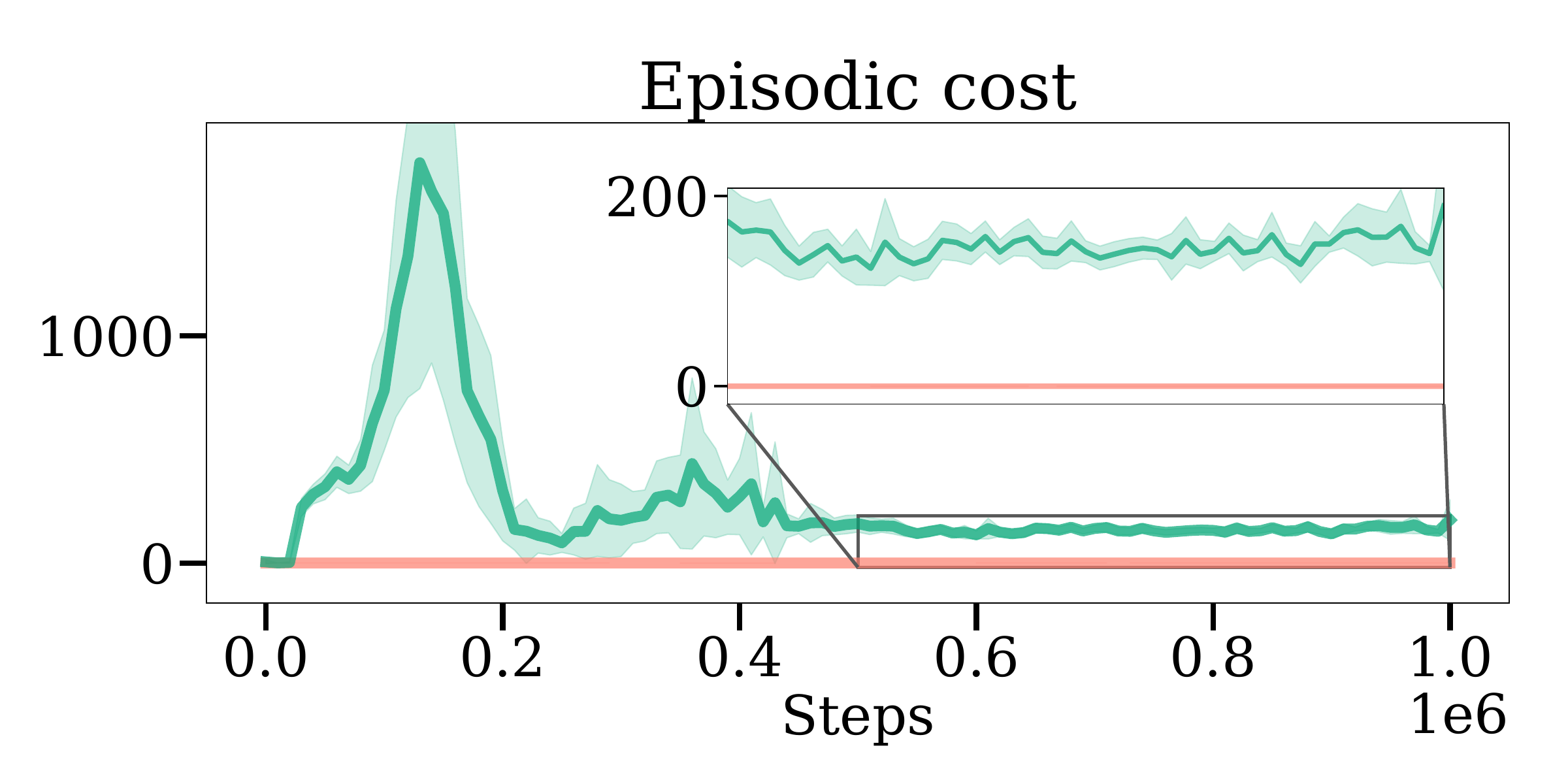}
        \caption{Quadrotor}
    \end{subfigure}
    \vspace{0.5em}
    
    \includegraphics[height=1.25em]{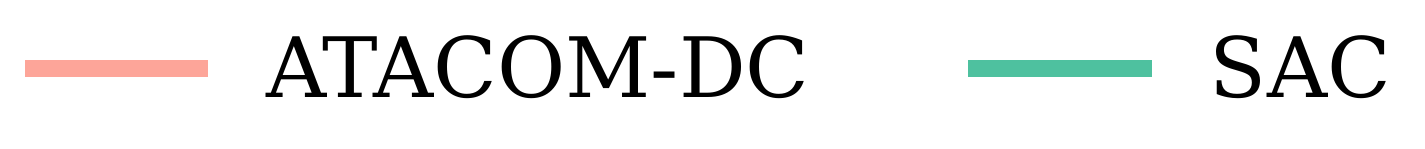}
    \caption{Comparison of our method against unconstrained SAC in the planar air hockey and quadrotor tasks, both in terms of task metric (top row) and safety (bottom row)}
    \label{fig:comparison_unconstrained}
    \vspace{-1.5em}
\end{figure*}

\paragraph*{Air hockey task}
In the air hockey task, a KUKA iiwa14 robotic manipulator is trained to strike a puck, initialized at random positions on the table, toward the opponent’s goal. The robot observes both its own state and the puck state, including joint configurations and puck motion. Based on these observations, the policy outputs desired joint velocities, which are executed by a low-level controller to generate smooth torque commands.
The objective of the task is to maximize performance by effectively hitting the puck, increasing its forward velocity, and ultimately scoring a goal. To ensure feasibility and safety, constraints are imposed on joint limits, workspace boundaries defined by multiple planes around the table, and collision avoidance by enforcing minimum height limits on selected robot links.

\paragraph*{Planar air hockey task}
In the planar air hockey task, the agent controls a 3-DoF planar robotic arm equipped with a mallet at the end-effector that aims to strike a puck toward the opponent’s goal using the same constrained setup as the air hockey task. This task can be controlled either in acceleration or in velocity.

\paragraph*{Quadrotor navigation task}
The quadrotor navigation task involves a quadrotor drone required to track a moving target that follows an eight-shaped trajectory, while avoiding a large cylindrical obstacle placed in the center of a confined environment. The control input consists of the total thrust generated by the propellers and the torques applied along the roll, pitch, and yaw axes. The agent observes the robot’s proprioceptive state and the target’s position and velocity.
To guarantee safe and stable navigation, several constraints are enforced. These include collision-avoidance constraints to prevent impacts with obstacles, workspace constraints that confine the quadrotor within boundaries defined along the $x$, $y$, and $z$ axes, and an additional constraint limiting the angular velocity around the $z$-axis to improve flight stability.

\subsection{Experimental setup}
The following results compare the proposed \gls{atacomdc} with the unconstrained approach, the original \gls{atacom}, and the \gls{datacom} method. Each experiment is repeated over 15 independent random seeds.
All approaches employ SAC~\cite{haarnoja2018soft} as the underlying \gls{rl} policy optimization algorithm. 
The policy is an MLP with $2$ layers parameterizing a state-dependent Gaussian distribution over actions.
While we log the cumulative discounted return, for presentation reasons we only show task-specific performance metrics. 
For the air hockey setups (both KUKA and planar), these include the goal success rate and the puck velocity. 
For the quadrotor task, the performance metric is the distance from the moving target.
For safety evaluation, we consider the \emph{episodic cost}, defined as the
cumulative cost over an episode, where the cost at timestep $t$ is $max(k(s_t), 0)$, as in~\cite{guenster2024handling}.

\subsection{Performance against unconstrained methods}
In Figure~\ref{fig:comparison_unconstrained}, we present a comparison of our approach in all the tasks against the unconstrained \gls{sac} algorithm. In this setting, we do not report results for the kuka iiwa air hockey task, as \gls{sac} is unable to learn in such a complex setup without the possibility of exploiting the constraint information.
As a result of this experiment, we demonstrate that the optimal policy is safe for some tasks, and is not safe for others. In fact, in quadrotor and planar air hockey controlled in acceleration, the \gls{sac} final policy still violates constraints, thus achieving higher task performance.
Instead, \gls{atacomdc} maintains safety throughout the entire training process, with similar final performance and generally better or comparable learning curves. 

\subsection{Impact of Directional Constraints}
\begin{figure*}[t]
    \centering
    \begin{subfigure}[t]{0.29\textwidth}
        \centering
        \includegraphics[width=\textwidth]{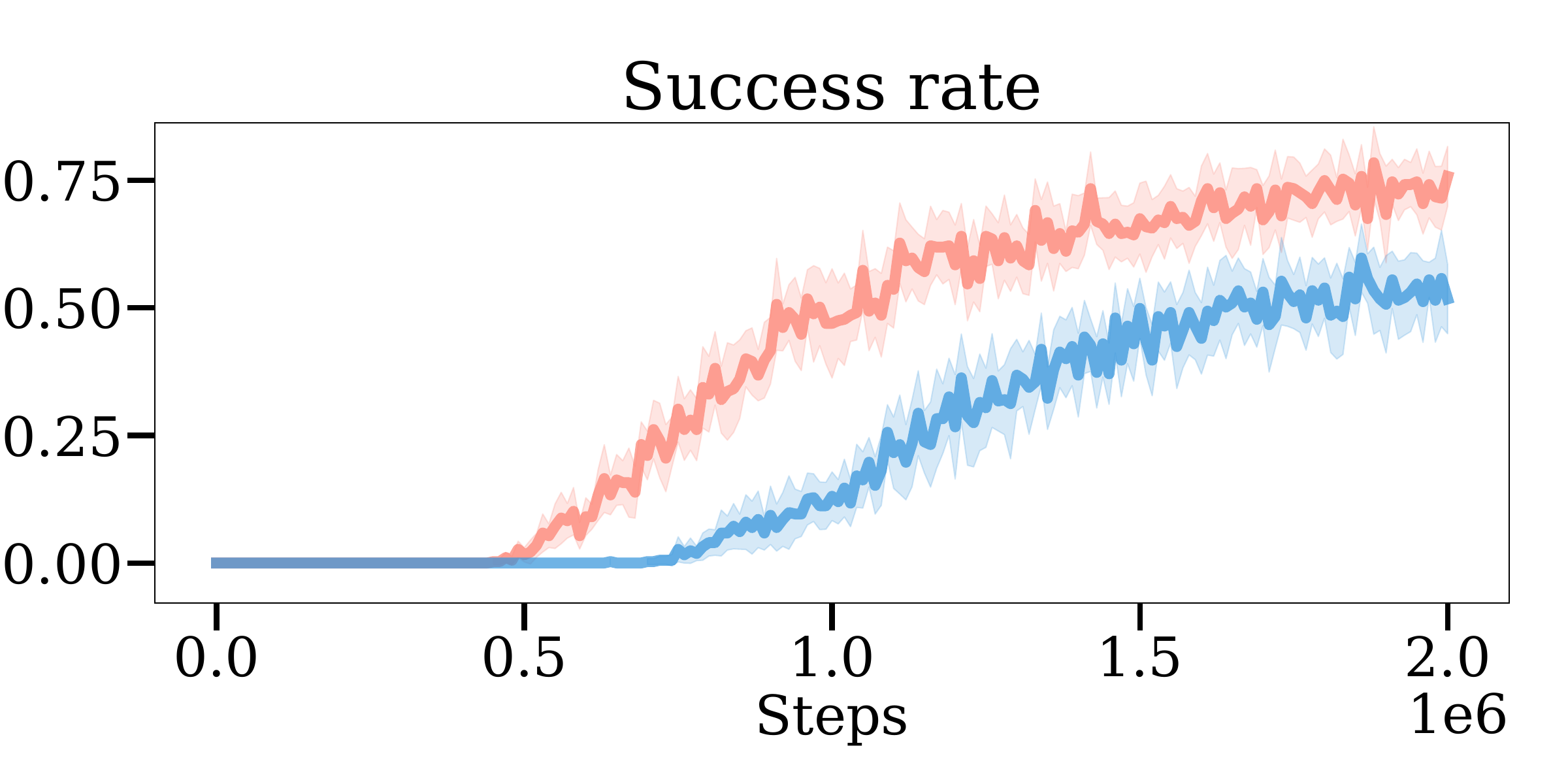}
    \end{subfigure}
    \begin{subfigure}[t]{0.29\textwidth}
        \centering
        \includegraphics[width=\textwidth]{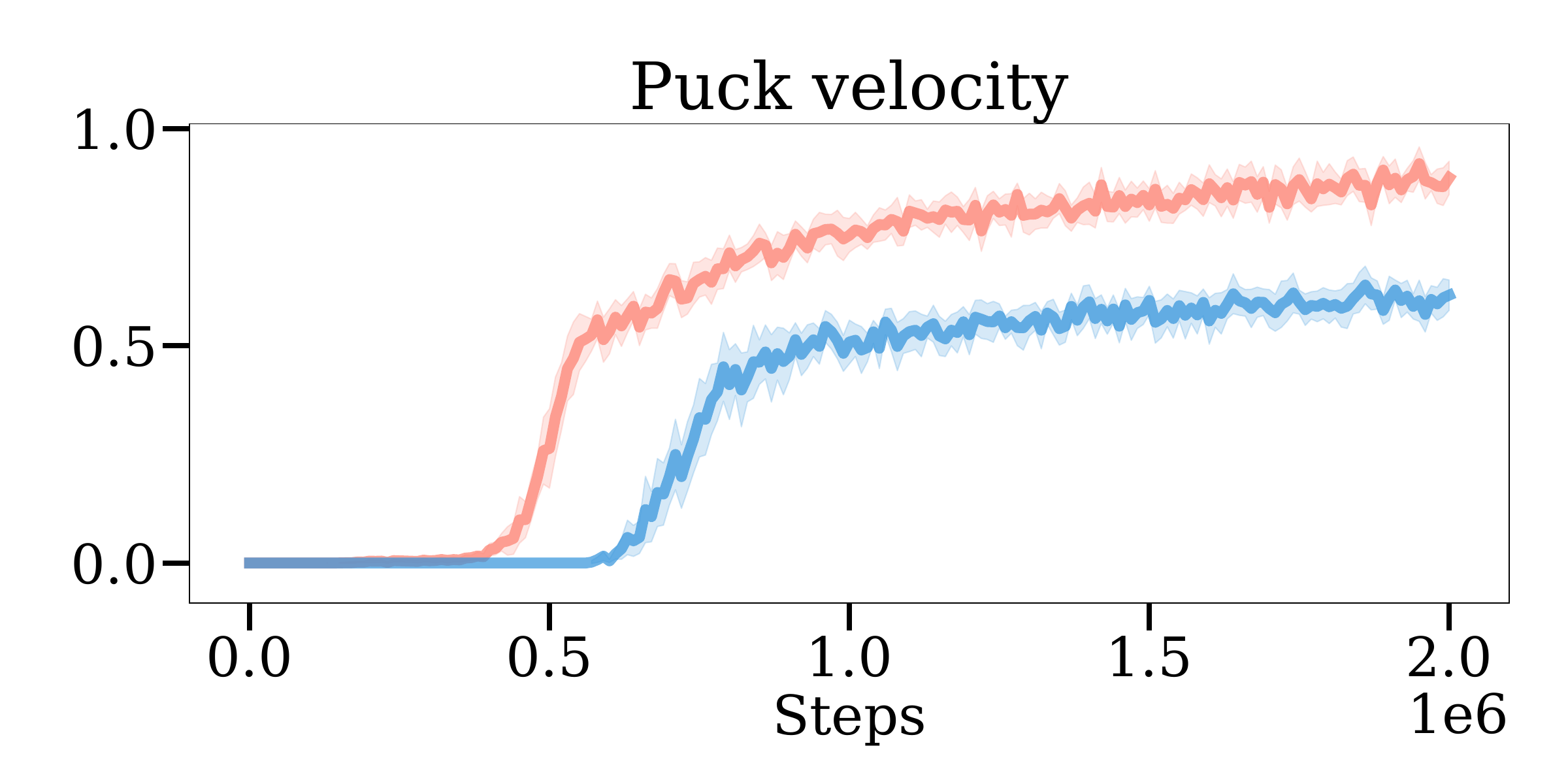}
    \end{subfigure}
    \begin{subfigure}[t]{0.29\textwidth}
        \centering
        \includegraphics[width=\textwidth]{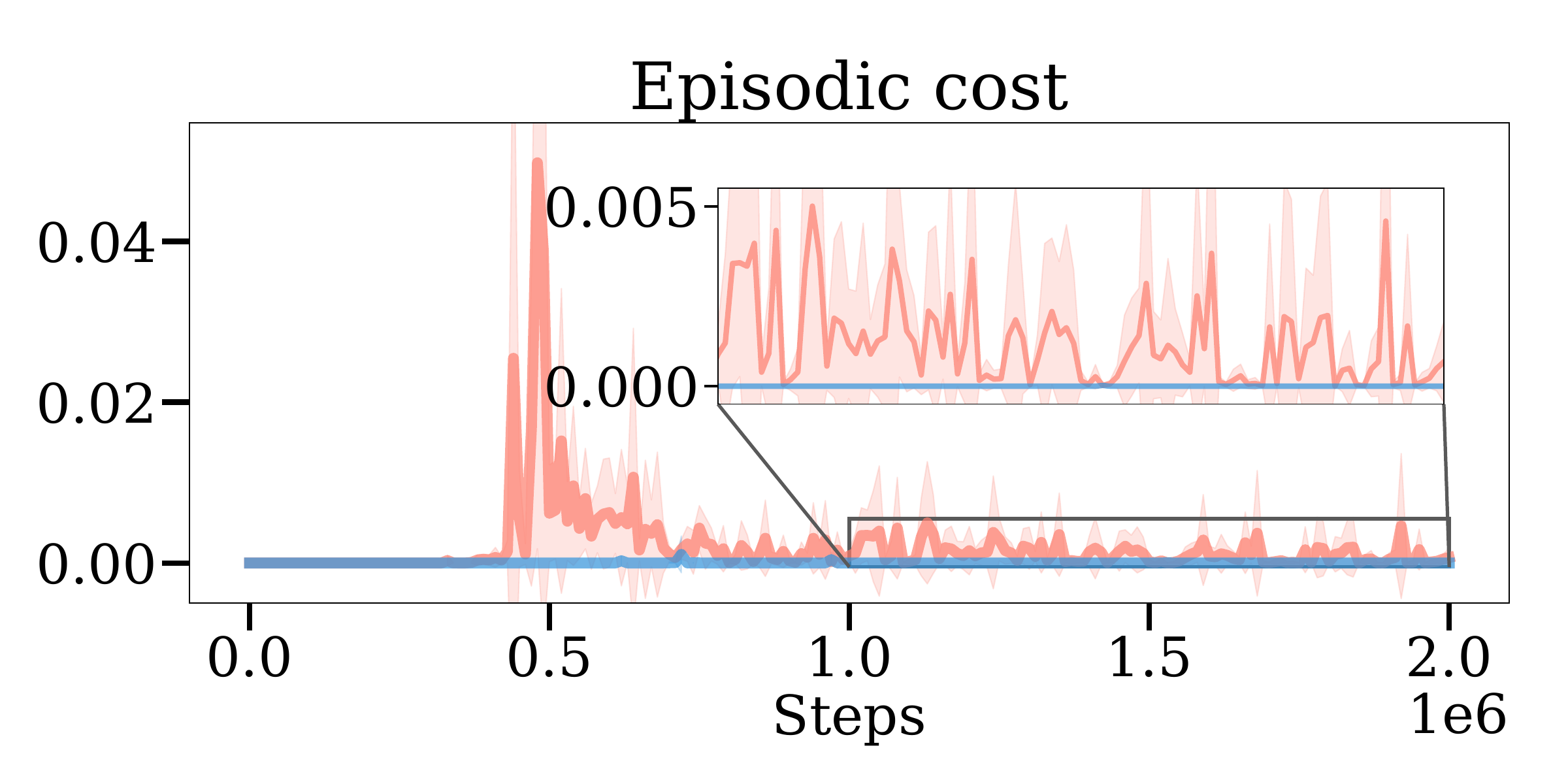}
    \end{subfigure}

    \includegraphics[height=1.25em]{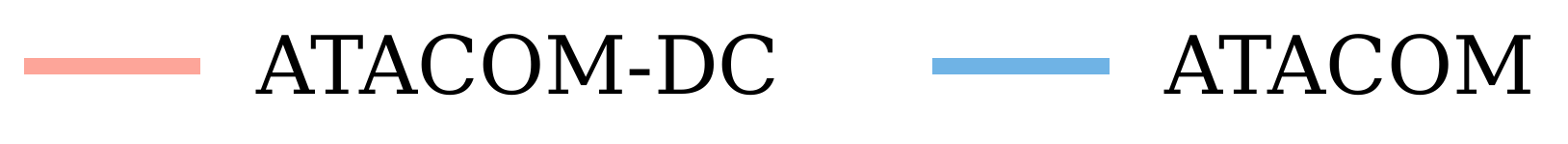}
    \vspace{-0.5em}
    
    \caption{Impact of directional constraints in the kuka iiwa air Hockey task}
    \label{fig:atacom_comparison_iiwa}
    \vspace{-1em}
\end{figure*}

\begin{figure*}[t]
    \centering
    \begin{subfigure}[t]{0.29\textwidth}
        \centering
        \includegraphics[width=\textwidth]{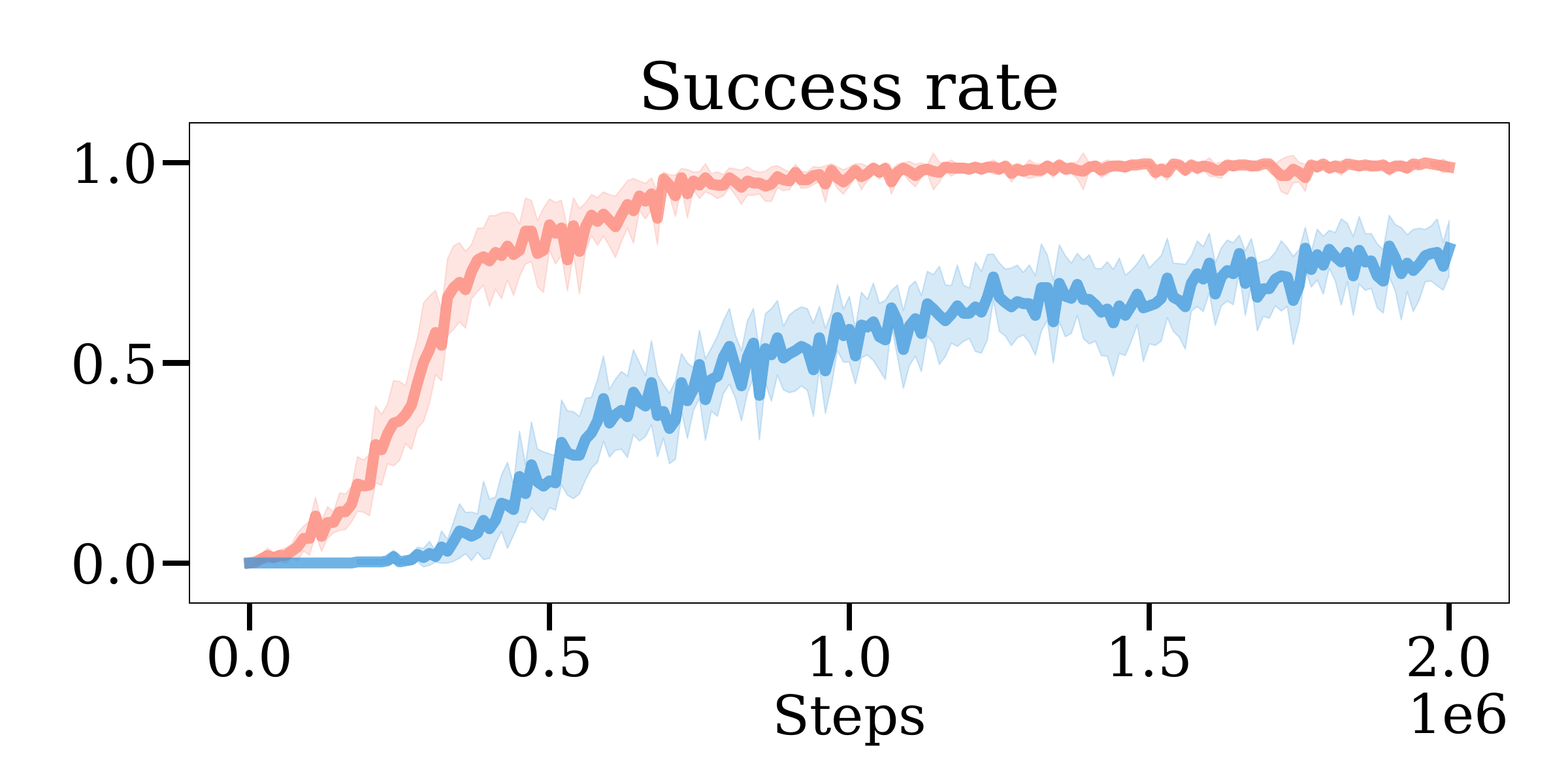}
    \end{subfigure}
    \begin{subfigure}[t]{0.29\textwidth}
        \centering
        \includegraphics[width=\textwidth]{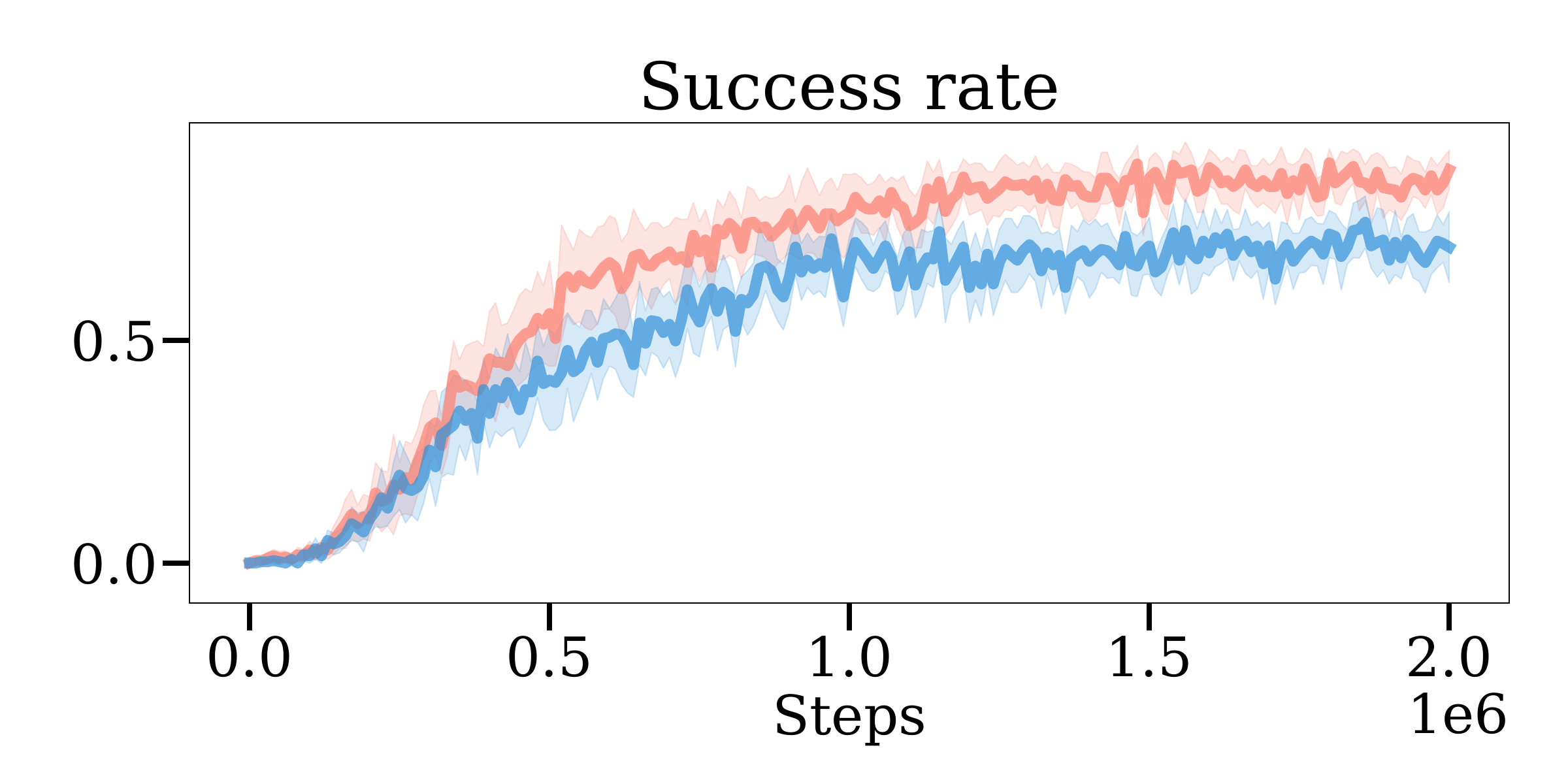}
    \end{subfigure}
    \begin{subfigure}[t]{0.29\textwidth}
        \centering
        \includegraphics[width=\textwidth]{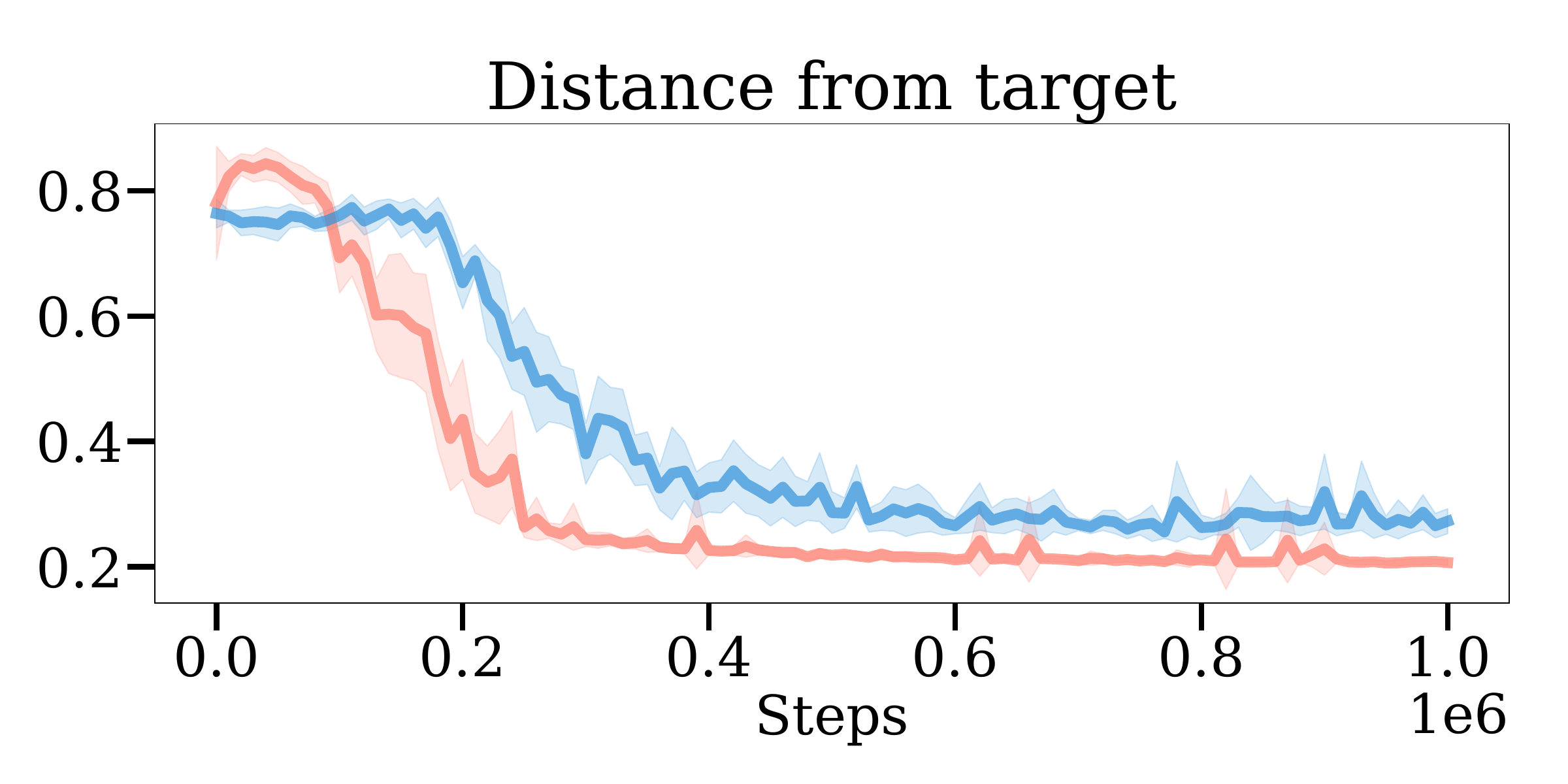}
        
    \end{subfigure}

    \begin{subfigure}[t]{0.29\textwidth}
        \centering
        \includegraphics[width=\textwidth]{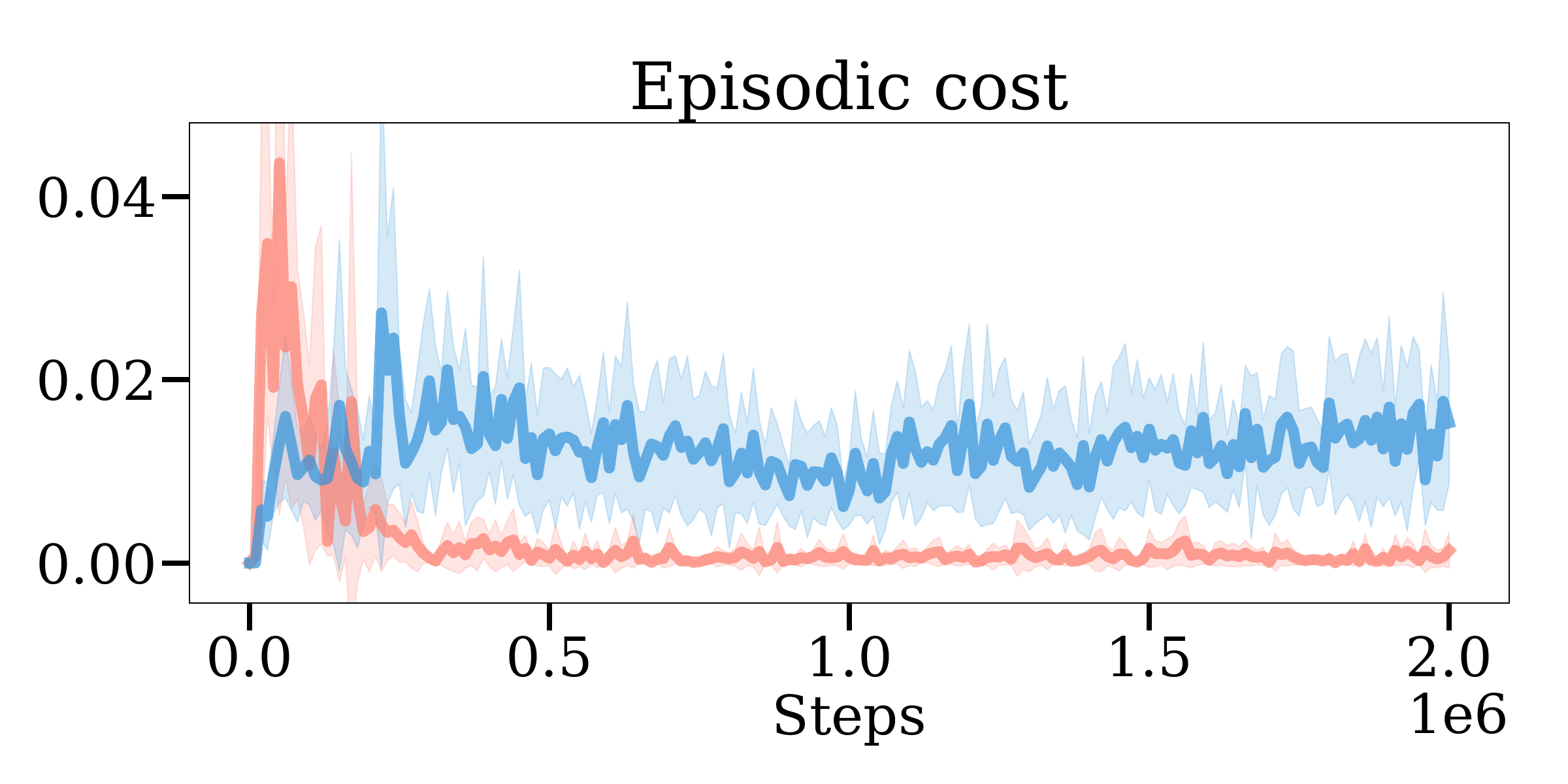}
        \caption{Planar air hockey (velocity)}
    \end{subfigure}
    \begin{subfigure}[t]{0.29\textwidth}
        \centering
        \includegraphics[width=\textwidth]{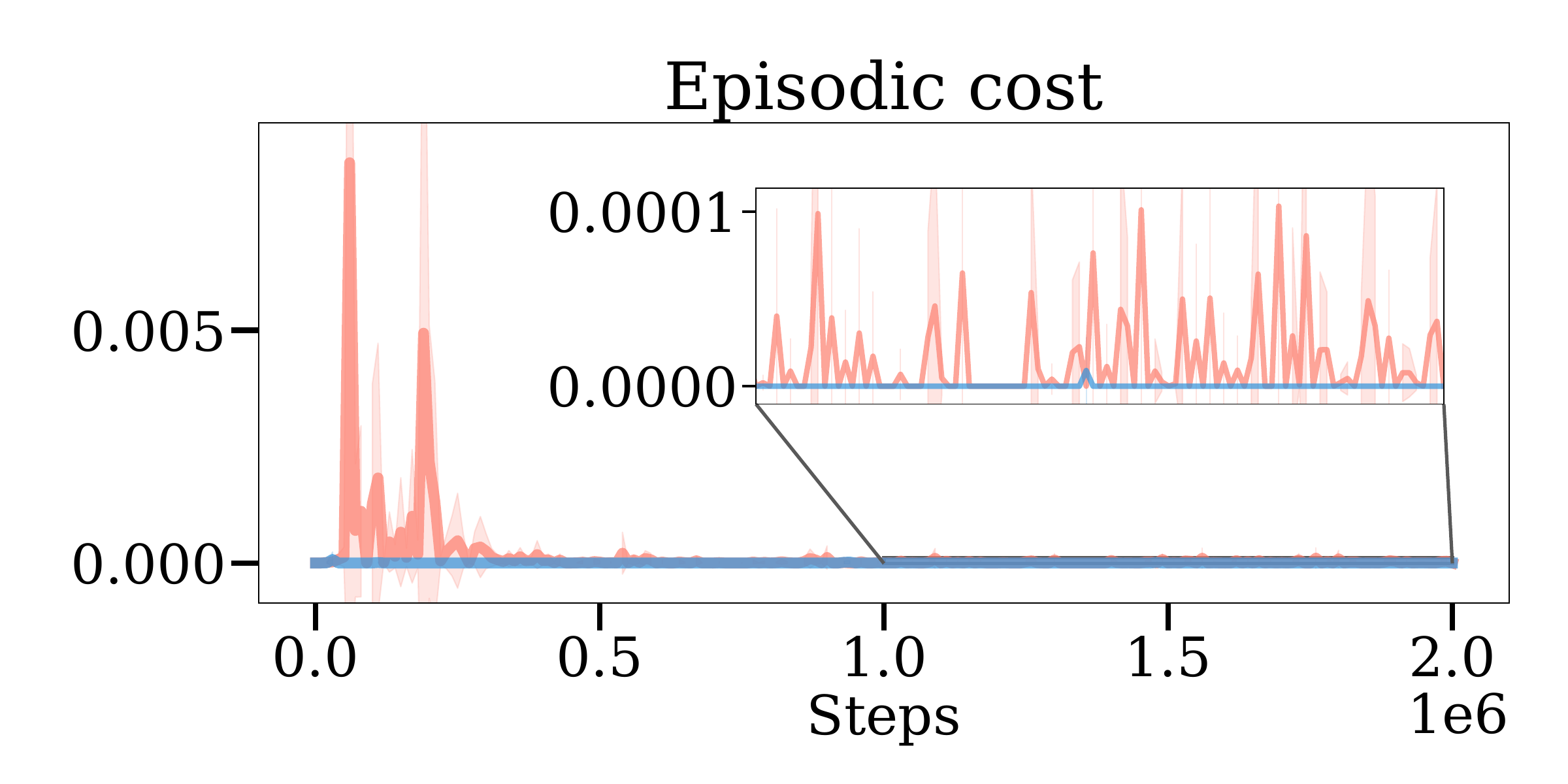}
        \caption{Planar air hockey (acceleration)}
    \end{subfigure}
    \begin{subfigure}[t]{0.29\textwidth}
        \centering
        \includegraphics[width=\textwidth]{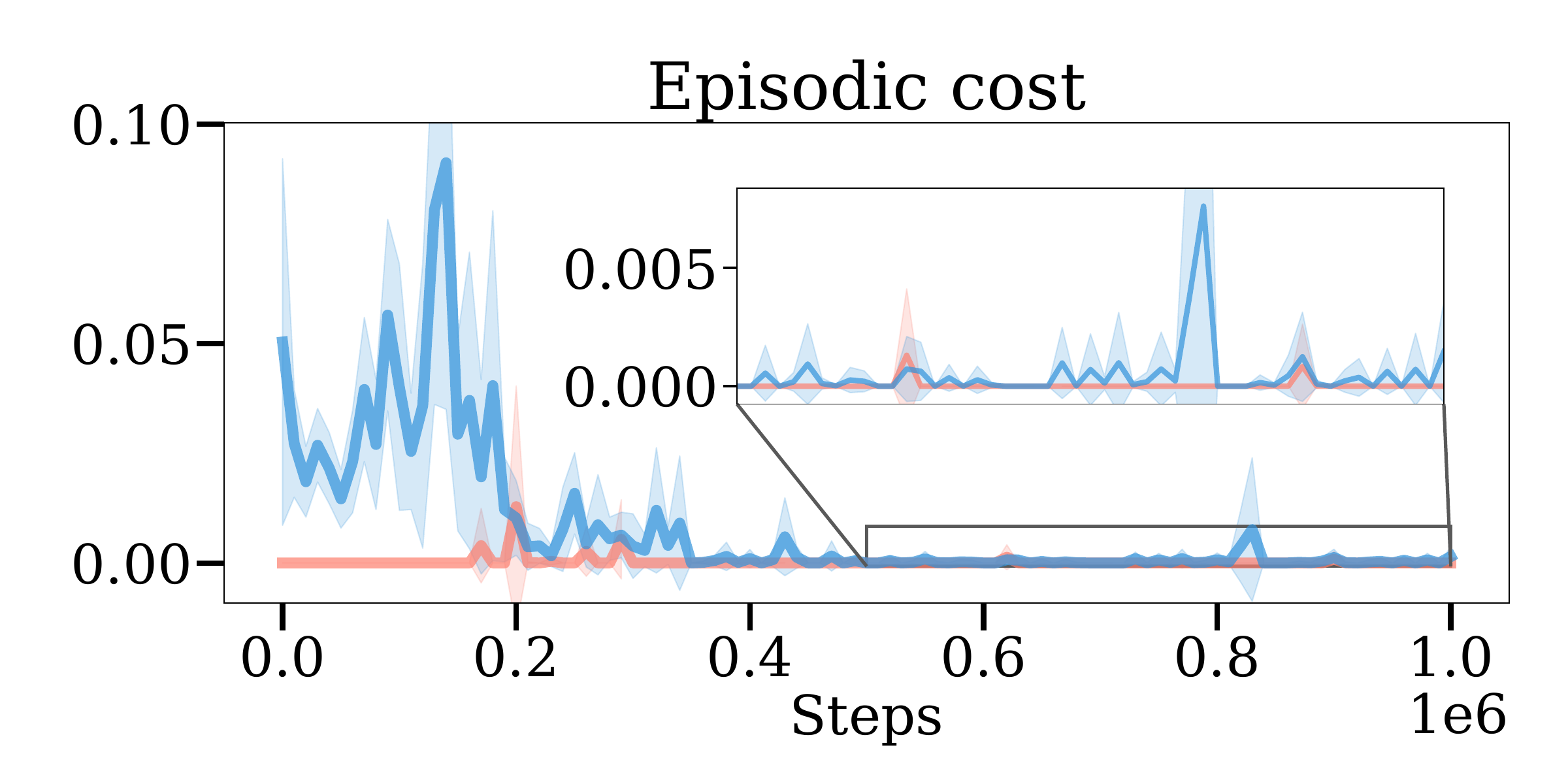}
        \caption{Quadrotor}
    \end{subfigure}

    \vspace{0.5em}
    \includegraphics[height=1.25em]{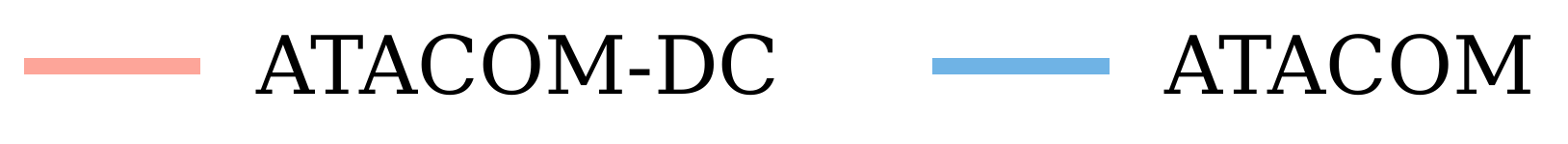}
    
    \caption{Impact of directional constraints in the planar air hockey tasks and in the quadrotor task. The figure presents task performance (top row) and the safety violations (bottom row).}
    \label{fig:atacom_comparison_others}
    \vspace{-1.5em}
\end{figure*}
In Figures~\ref{fig:atacom_comparison_iiwa} and~\ref{fig:atacom_comparison_others}, we analyze the impact of Directional Constraints against the vanilla \gls{atacom} algorithm. Our results show that in all tasks
Directional Constraints consistently lead to faster learning and overall improved final performance, while constraint violations are generally reduced across tasks or at least comparable, except during the first epochs. Indeed, Directional Constraints are less restrictive, allowing for more movement in the initial phases of learning.
However, when the task objective is not to push against the constraints, this behavior allows the system to learn to leave the unsafe area, leading to lower long-term constraint violations. In any case, all the approaches show very low, close to zero, constraint violations.

The benefit of enhanced exploration is particularly impactful in the air hockey setting, as highlighted in Figure~\ref{fig:atacom_comparison_iiwa}. In this complex scenario, the less restrictive exploration allows us to learn faster and achieve higher-speed and more precise policies, which results in higher success rate and puck velocities. However, the faster motion may cause small violations due to the model inaccuracies, as it does not include the torque model and the low-level controller.

\subsection{D-ATACOM Improvements}
\begin{figure*}[t]
    \centering
    \begin{minipage}{0.64\textwidth}
        \begin{subfigure}[t]{0.5\textwidth}
            \centering
            \includegraphics[width=\textwidth]{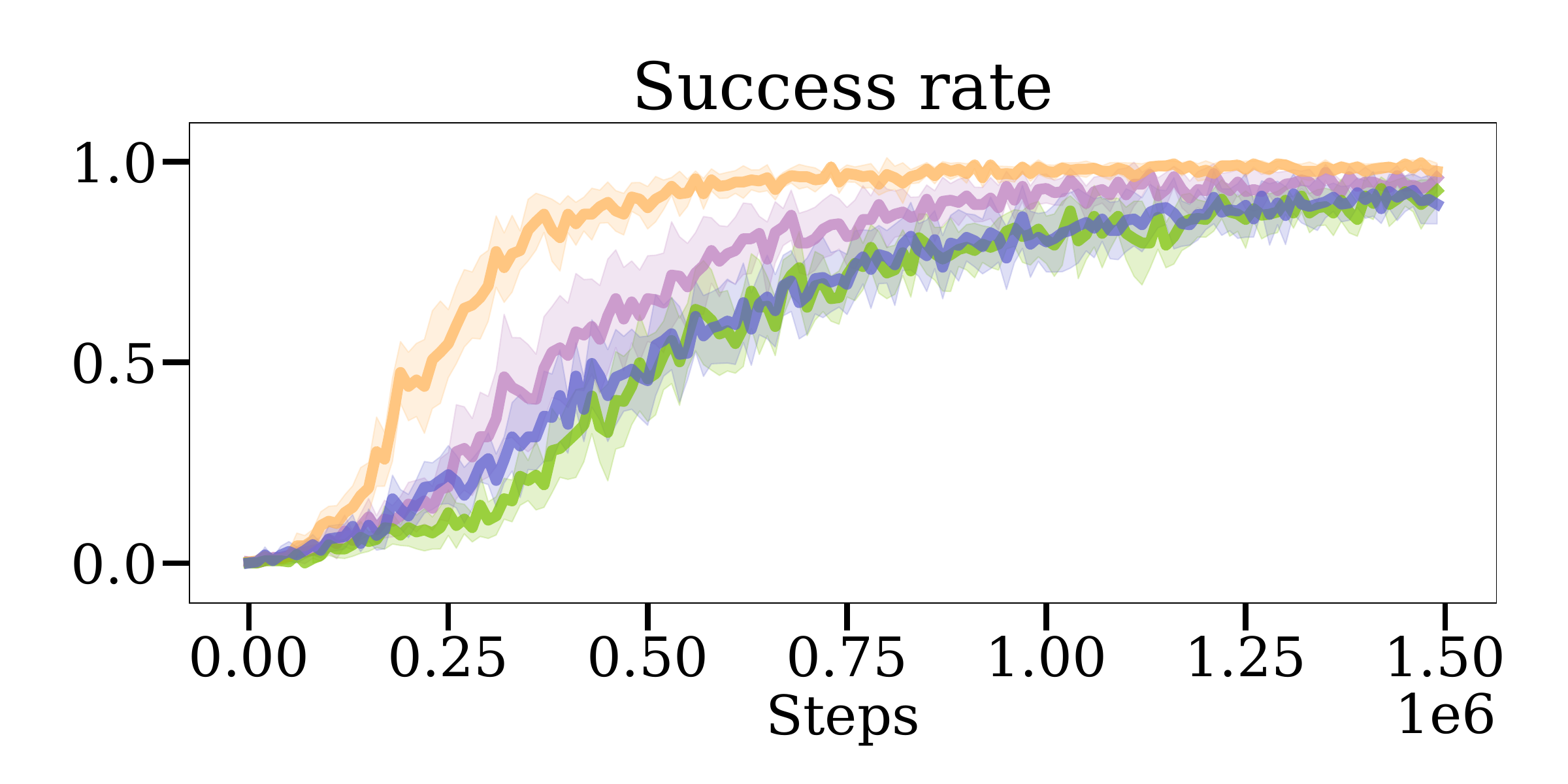}
        \end{subfigure}
        \begin{subfigure}[t]{0.5\textwidth}
            \centering
            \includegraphics[width=\textwidth]{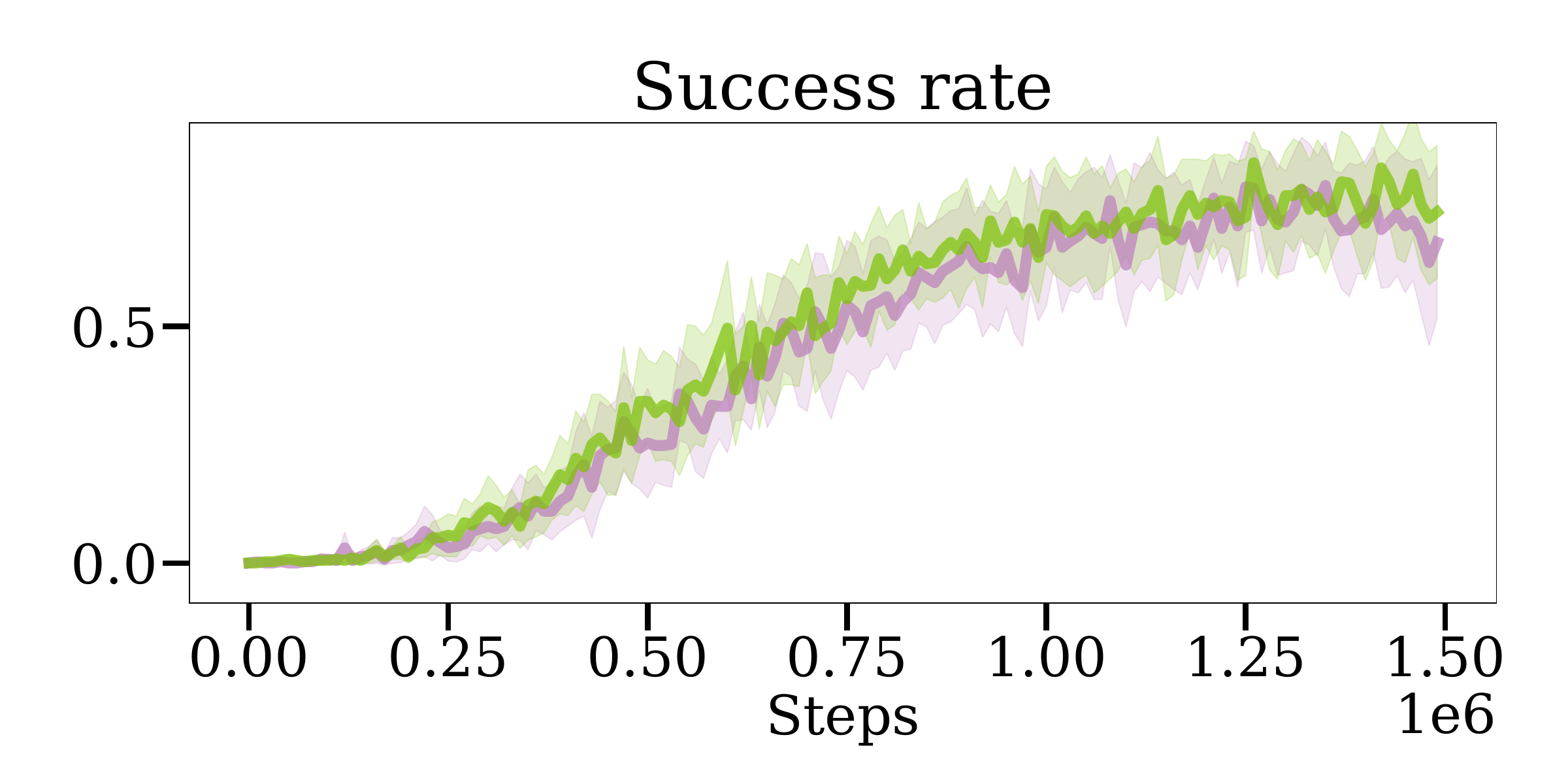}
        \end{subfigure}
        
        \begin{subfigure}[t]{0.5\textwidth}
            \centering
            \includegraphics[width=\textwidth]{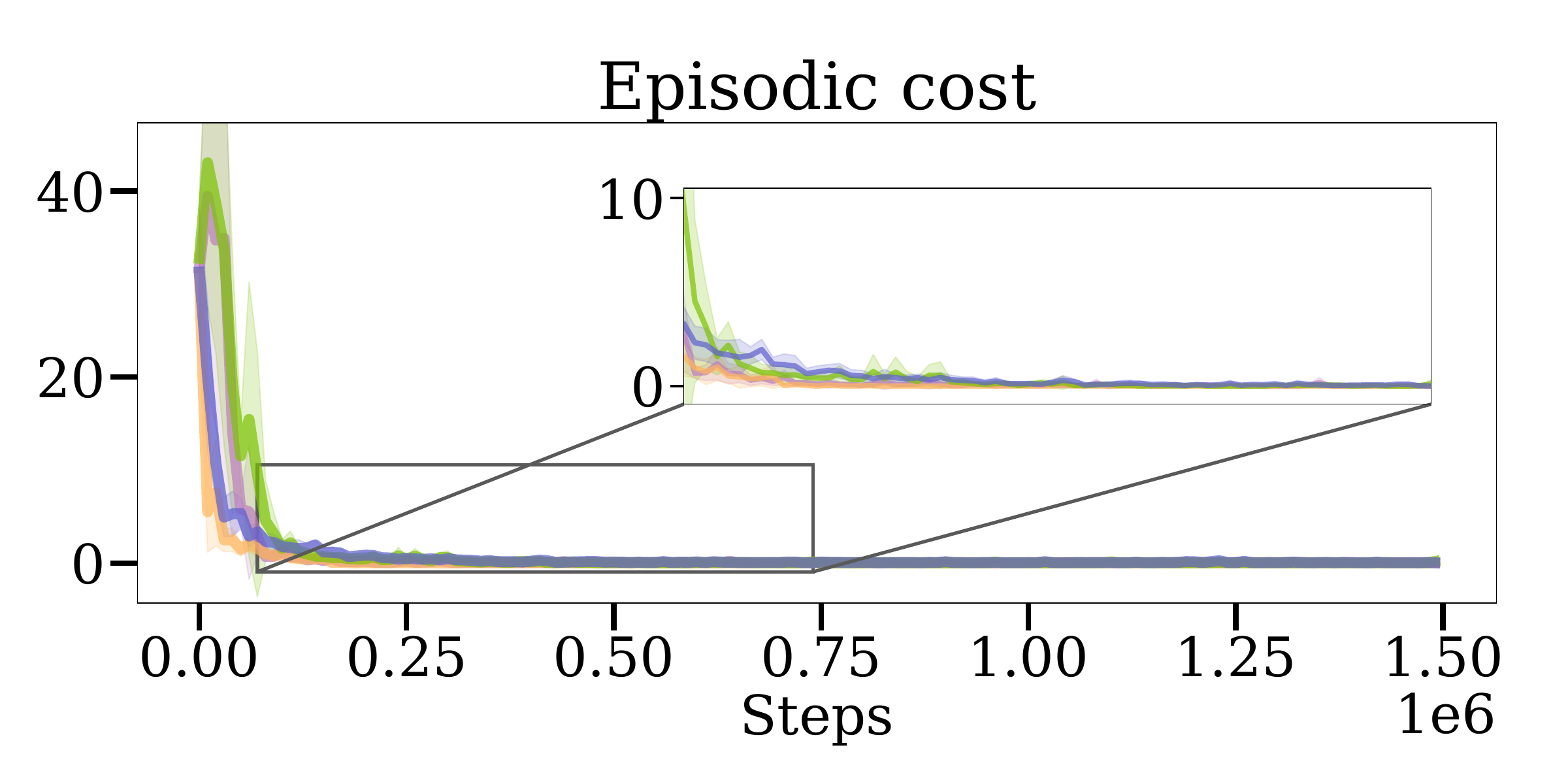}
            \caption{Planar air hockey (velocity)}
            \label{fig:datacom_constraints_comparison}
        \end{subfigure}
        \begin{subfigure}[t]{0.5\textwidth}
            \centering
            \includegraphics[width=\textwidth]{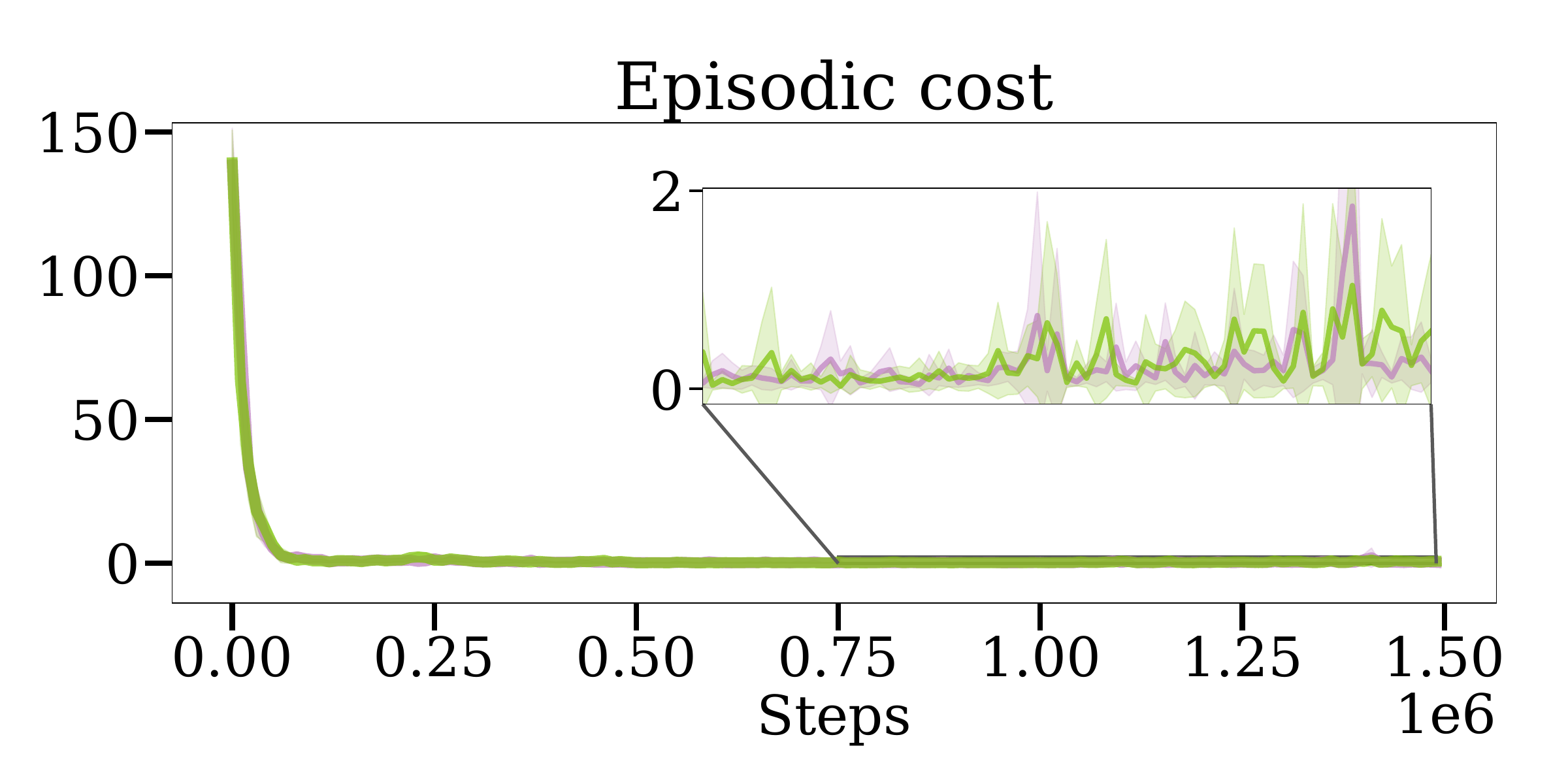}
            \caption{Planar air hockey (acceleration)}
        \end{subfigure}
        \begin{minipage}{0.29\textwidth}
            
        \end{minipage}
    \end{minipage}
    \begin{subfigure}[c]{0.3\textwidth}
        \centering
        \includegraphics[width=\columnwidth]{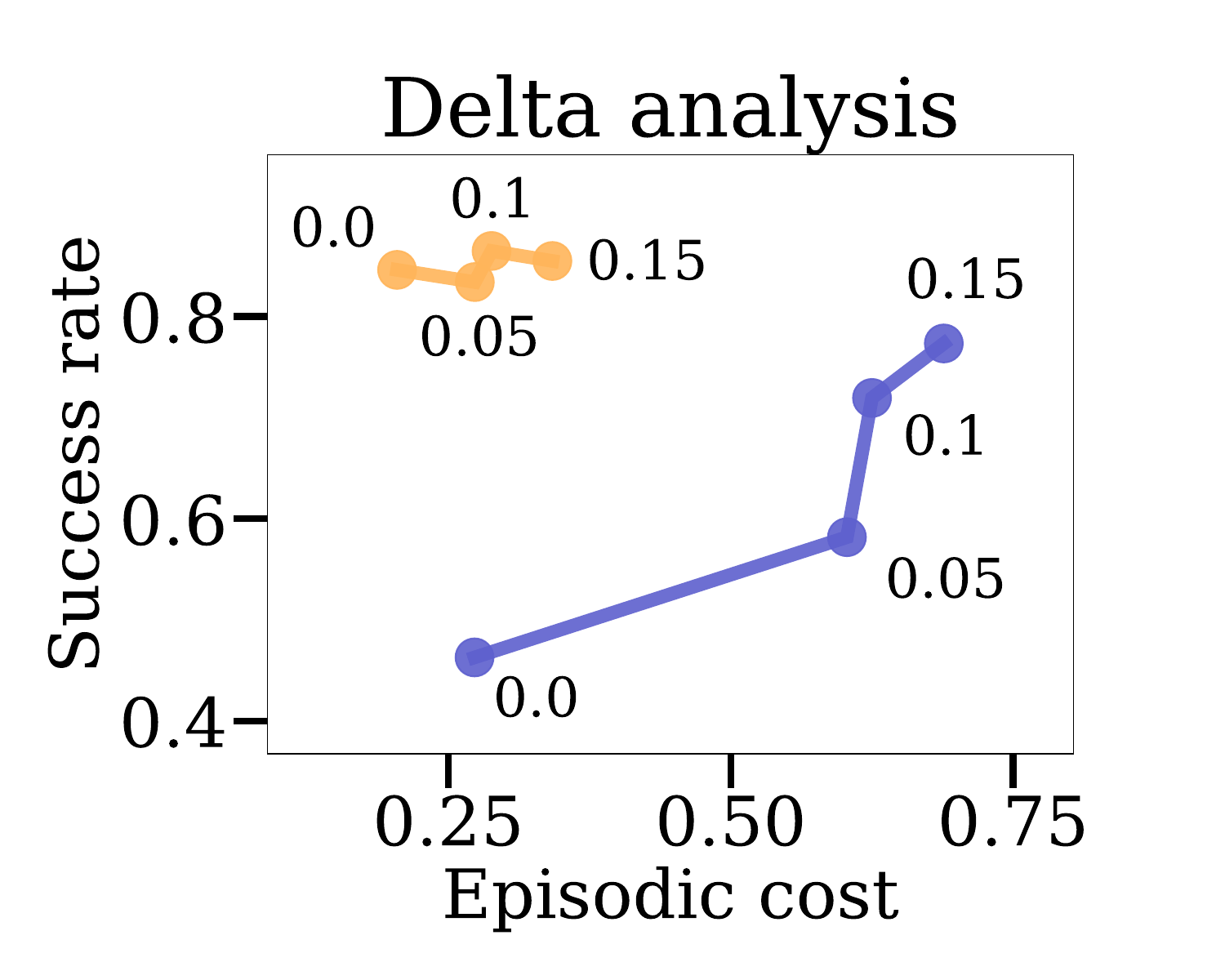}
        
        \caption{Planar air hockey - $\delta$ analysis}
        \label{fig:datacom_constraints_delta}
    \end{subfigure}

    \vspace{0.5em}
        \includegraphics[height=1.25em]{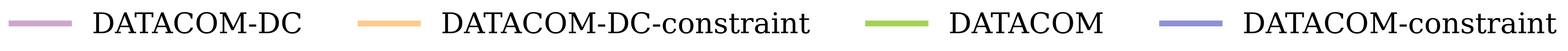}
    
    \caption{Impact of Directional constraints in the \gls{datacom} algorithm. The figure presents task performance (top row) and the safety violations (bottom row). The last column presents the results for direct constraint learning.}
    \label{fig:datacom_comparison}
    \vspace{-1.5em}
\end{figure*}

The methodology can be easily extended to the \gls{datacom} approach. Here, we report results only for the planar air hockey environment, both in velocity and acceleration.
In the planar velocity scenario, the result shows that, while having comparable performance in terms of safety as the vanilla \gls{datacom}, directional constraints boost the overall learning performances. However, directional constraints have no clear statistical impact in the planar air hockey acceleration scenario.

Furthermore, we investigate whether it is possible to learn without using a \gls{fvf}, only using the constraints and their uncertainty. Indeed, learning a constraint function is more stable and faster than learning a \gls{fvf}, which requires temporal difference learning. We investigate this in the setting of planar air hockey in velocity, which is a first-order system, where vanilla constraints are sufficient to impose safety to the system.

Unfortunately, directly learning the constraints and avoiding learning the \gls{fvf} produces comparable performance, at the cost of increased constraint violations. Notably, if we also fix the dynamic threshold $\delta$, tuned automatically by \gls{datacom} to trade off exploration and exploitation, and remove warm-up trajectories, we can achieve a performance boost in terms of learning speed and safety, as clearly shown in Figure.~\ref{fig:datacom_constraints_comparison}, where the DATACOM-DC-constraint line, which represents this version of the algorithm, outperforms all other approaches. In particular, the safety improvement is mostly due to the reduced unconstrained warm-up phase, which limits early unsafe exploration while still allowing the agent to learn the constraints. This is due both to the removal of warm-up trajectories and the faster convergence to the desired constraint, allowed by the lack of temporal difference learning.

Furthermore, we analyze the sensitivity of the effect of the $\delta$ parameter over 5 different seeds for each value, comparing D-ATACOM-DC with \gls{datacom}. Here, having a higher $\delta$ results in higher constraint violations. However, looser constraints allow for more exploration, particularly in the initial episodes, when the constraint is not yet correctly approximated. Our results, presented in Figure~\ref{fig:datacom_constraints_delta}, show that the Directional Constraints yield a higher success rate, while keeping the cost lower for all values of $\delta$, showing that the proposed method is Pareto-optimal w.r.t. the baseline.

\subsection{Beta Analysis: Performance–Safety Trade-off}
\begin{figure}[t]
    
    \centering
    \begin{subfigure}[t]{0.49\columnwidth}
        \centering
        \includegraphics[width=\columnwidth]{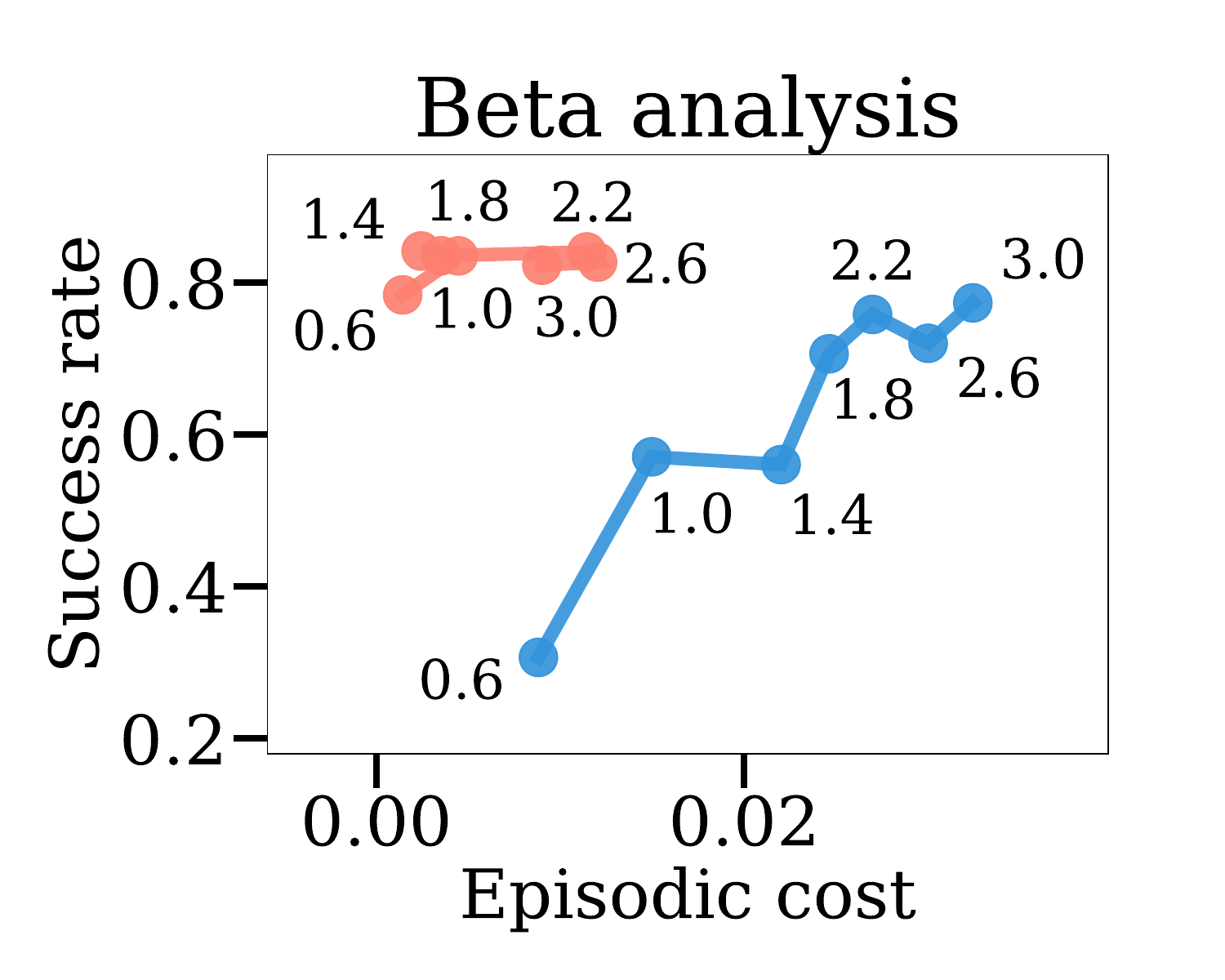}

        \caption{Planar air hockey}
    \end{subfigure}
    \begin{subfigure}[t]{0.49\columnwidth}
        \centering
        \includegraphics[width=\columnwidth]{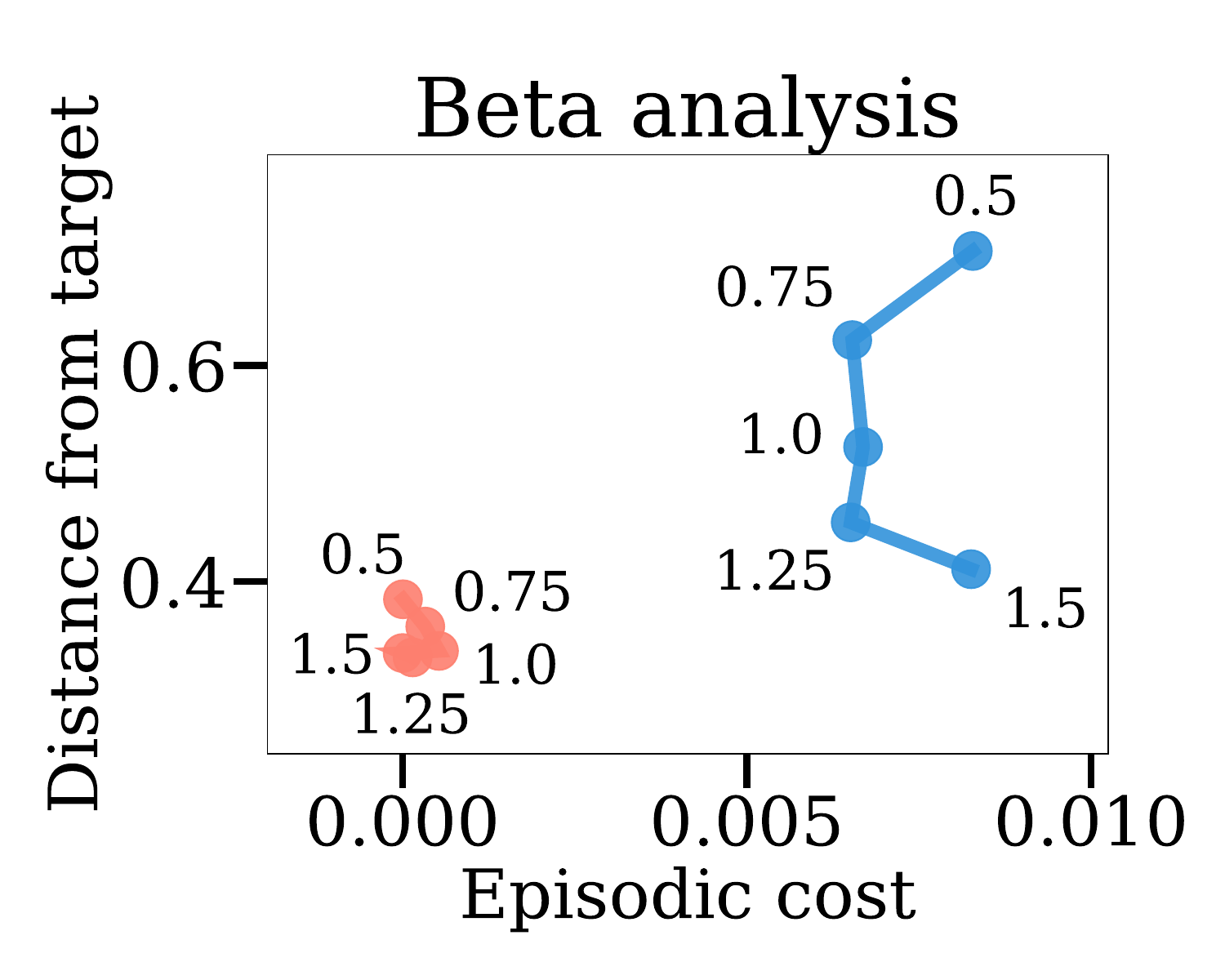}
        \caption{Quadrotor}
    \end{subfigure}
    \vspace{0.5em}
    
    \includegraphics[height=1.25em]{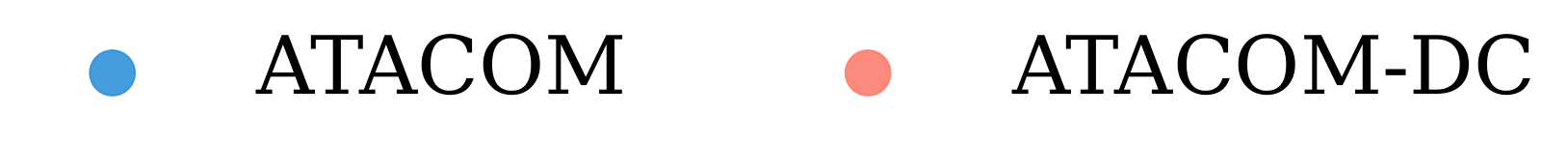}
    \caption{Analysis of the effects of the $\beta$ parameter in the planar air hockey and quadrotor tasks}
    \label{fig:hp_analysis}
    \vspace{-1.5em}
\end{figure}
Finally, we perform an ablation on the role of the $\beta$ parameter of the original \gls{atacom} safety layer for slack dynamics. Each value is evaluated across 5 independent seeds.
Results for different betas are reported in Figure~\ref{fig:hp_analysis} for both the planar air hockey and the quadrotor tasks. Results show that Directional Constraints allow effective operation even when a larger safety margin from the constraints is required, without degrading performance. In practice, \gls{atacomdc} improves the performance–safety trade-off, resulting in a Pareto-superior behavior compared to the baseline.







\section{Conclusion}
\label{sec:conclusion}
In this paper, we introduced \gls{atacomdc}, an improved version of the \gls{atacom} safety layer that boosts exploration capabilities simply and effectively through the concept of \emph{Directional Constraints}. The approach only modifies safe actions, retaining the safety guarantees of the original framework. Our simulated experiments show that the learning speed is increased while not compromising safety. On the contrary, in some settings, the safety guarantees are improved, as the agent policy can move away from the constraint quickly, moving the policy state distribution towards safer states.
Using Directional Constraints, in settings where the long-term safety is not a concern, it is not necessary to use TD-learning and automatic tuning of safety margin, allowing the algorithm to directly learn a constraint, which is faster and more accurate than learning a \gls{fvf}.
Furthermore, we show that the new approach is less sensitive to the safety parameters and allows us to obtain a better performance-safety trade-off, which is Pareto-dominant compared to vanilla \gls{atacom}, at least in our experimental setting.

In future work, we aim to bring these exploration advances together with more modern off-policy actor-critic algorithms to rapidly learn from scratch in complex real-world environments, e.g., the Robot Air Hockey, overcoming the limitations of the previous approach.

\bibliographystyle{IEEEtran}
\bibliography{bib} 

@inproceedings{rudin2022learning,
  title={Learning to walk in minutes using massively parallel deep reinforcement learning},
  author={Rudin, Nikita and Hoeller, David and Reist, Philipp and Hutter, Marco},
  booktitle={Conference on robot learning},
  pages={91--100},
  year={2022},
  organization={PMLR}
}

@inproceedings{zhuang2023robot,
  author    = {Zhuang, Ziwen and Fu, Zipeng and Wang, Jianren and Atkeson, Christopher and Schwertfeger, Sören and Finn, Chelsea and Zhao, Hang},
  title     = {Robot Parkour Learning},
  booktitle = {Conference on Robot Learning ({CoRL})},
  year      = {2023},
}

@inproceedings{huang2023dynamic,
    title={Dynamic Handover: Throw and Catch with Bimanual Hands},
    author={Binghao Huang and Yuanpei Chen and Tianyu Wang and Yuzhe Qin and Yaodong Yang and Nikolay Atanasov and Xiaolong Wang},
    booktitle={7th Annual Conference on Robot Learning},
    year={2023}
}

@inproceedings{lin2024twisting,
    title={Twisting Lids Off with Two Hands},
    author={Toru Lin and Zhao-Heng Yin and Haozhi Qi and Pieter Abbeel and Jitendra Malik},
    booktitle={8th Annual Conference on Robot Learning},
    year={2024},
    url={https://openreview.net/forum?id=3wBqoPfoeJ}
}

@inproceedings{li2025morphologically,
    title={Morphologically Symmetric Reinforcement Learning for Ambidextrous Bimanual Manipulation},
    author={Zechu Li and Yufeng Jin and Daniel Ordonez-Apraez and Claudio Semini and Puze Liu and Georgia Chalvatzaki},
    booktitle={9th Annual Conference on Robot Learning},
    year={2025}
}

@inproceedings{tobin2017domain,
  title={Domain randomization for transferring deep neural networks from simulation to the real world},
  author={Tobin, Josh and Fong, Rachel and Ray, Alex and Schneider, Jonas and Zaremba, Wojciech and Abbeel, Pieter},
  booktitle={2017 IEEE/RSJ international conference on intelligent robots and systems (IROS)},
  pages={23--30},
  year={2017},
  organization={IEEE}
}

@article{altman1998constrained,
  author  = {Altman, Eitan},
  journal = {Mathematical methods of operations research},
  number  = {3},
  pages   = {387--417},
  title   = {{Constrained Markov Decision Processes with Total Cost Criteria: Lagrangian Approach and Dual Linear Program}},
  volume  = {48},
  year    = {1998}
}

@inproceedings{achiam2017constrained,
    author = {Achiam, Joshua and Held, David and Tamar, Aviv and Abbeel, Pieter},
    booktitle = {International Conference on Machine Learning (ICML)},
    title = {{Constrained Policy Optimization}},
    year = {2017}
}

@article{ray2019benchmarking,
  title={Benchmarking safe exploration in deep reinforcement learning.(2019)},
  author={Ray, Alex and Achiam, Joshua and Amodei, Dario},
  journal={URL https://cdn. openai. com/safexp-short. pdf},
  volume={},
  pages={1--25},
  year={2019}
}

@inproceedings{liu2020ipo,
  title={Ipo: Interior-point policy optimization under constraints},
  author={Liu, Yongshuai and Ding, Jiaxin and Liu, Xin},
  booktitle={AAAI Conference on Artificial Intelligence (AAAI)},
  volume={34(04)},
  pages={4940--4947},
  year={2020}
}

@inproceedings{ding2021provably,
    author = {Ding, Dongsheng and Wei, Xiaohan and Yang, Zhuoran and Wang, Zhaoran and Mihailo R. Jovanovic},
    booktitle = {International Conference on Artificial Intelligence and Statistics (AISTATS)},
    title = {{Provably Efficient Safe Exploration via Primal-Dual Policy Optimization}},
    volume = {130},
    year = {2021}
}

@InProceedings{ha2021learning,
  title = {Learning to Walk in the Real World with Minimal Human Effort},
  author = {Ha, Sehoon and Xu, Peng and Tan, Zhenyu and Levine, Sergey and Tan, Jie},
  booktitle = {Proceedings of the 2020 Conference on Robot Learning},
  pages = {1110--1120},
  year = {2021},
  volume = {155},
  series = {Proceedings of Machine Learning Research},
  month = {16--18 Nov},
  publisher = {PMLR},
}

@inproceedings{yang2021wcsac,
  title={WCSAC: Worst-case soft actor critic for safety-constrained reinforcement learning},
  author={Yang, Qisong and Sim{\~a}o, Thiago D and Tindemans, Simon H and Spaan, Matthijs TJ},
  booktitle={Proceedings of the AAAI Conference on Artificial Intelligence},
  volume={35 (12)},
  pages={10639--10646},
  year={2021}
}

@article{yang2023safety,
  title={Safety-constrained reinforcement learning with a distributional safety critic},
  author={Yang, Qisong and Simao, Thiago D and Tindemans, Simon H and Spaan, Matthijs TJ},
  journal={Machine Learning},
  volume={112},
  number={3},
  pages={859--887},
  year={2023},
  publisher={Springer}
}

@article{yang2023feasible,
  title={Feasible Policy Iteration},
  author={Yang, Yujie and Zheng, Zhilong and Li, Shengbo Eben},
  journal={arXiv preprint arXiv:2304.08845},
  year={2023}
}

@inproceedings{liu2022robot,
  title={Robot reinforcement learning on the constraint manifold},
  author={Liu, Puze and Tateo, Davide and Ammar, Haitham Bou and Peters, Jan},
  booktitle={5th Conference on Robot Learning (CoRL)},
  year={2021},
  organization={PMLR}
}

@article{liu2025safe,
  title={Safe reinforcement learning on the constraint manifold: Theory and applications},
  author={Liu, Puze and Bou-Ammar, Haitham and Peters, Jan and Tateo, Davide},
  journal={IEEE Transactions on Robotics},
  year={2025},
  publisher={IEEE}
}

@inproceedings{liu2023safe,
  title={Safe reinforcement learning of dynamic high-dimensional robotic tasks: navigation, manipulation, interaction},
  author={Liu, Puze and Zhang, Kuo and Tateo, Davide and Jauhri, Snehal and Hu, Zhiyuan and Peters, Jan and Chalvatzaki, Georgia},
  booktitle={2023 IEEE International Conference on Robotics and Automation (ICRA)},
  pages={9449--9456},
  year={2023},
  organization={IEEE}
}

@inproceedings{guenster2024handling,
    title={Handling Long-Term Safety and Uncertainty in Safe Reinforcement Learning},
    author={Jonas G{\"u}nster and Puze Liu and Jan Peters and Davide Tateo},
    booktitle={Conference on Robot Learning ({CoRL})},
    year={2024},
}

@inproceedings{ames2019control,
  title={Control barrier functions: Theory and applications},
  author={Ames, Aaron D and Coogan, Samuel and Egerstedt, Magnus and Notomista, Gennaro and Sreenath, Koushil and Tabuada, Paulo},
  booktitle={2019 18th European control conference (ECC)},
  pages={3420--3431},
  year={2019},
  organization={IEEE}
}

@inproceedings{taylor2020learning,
  title={Learning for safety-critical control with control barrier functions},
  author={Taylor, Andrew and Singletary, Andrew and Yue, Yisong and Ames, Aaron},
  booktitle={Learning for Dynamics and Control},
  pages={708--717},
  year={2020},
  organization={PMLR}
}

@article{xiao2022high_order,
  title    = {High-{Order} {Control} {Barrier} {Functions}},
  volume   = {67},
  issn     = {1558-2523},
  doi      = {10.1109/TAC.2021.3105491},
  number   = {7},
  journal  = {IEEE Transactions on Automatic Control},
  author   = {Xiao, Wei and Belta, Calin},
  year     = {2022},
  pages    = {3655--3662},
}

@article{tan2023your,
  title={Your Value Function is a Control Barrier Function: Verification of Learned Policies using Control Theory},
  author={Tan, Daniel CH and Acero, Fernando and McCarthy, Robert and Kanoulas, Dimitrios and Li, Zhibin Alex},
  journal={2nd Workshop on Formal Verification of Machine Learning in the 40th International Conference on Machine Learning},
  year={2023}
}

@article{yang2023model,
  title={Model-free safe reinforcement learning through neural barrier certificate},
  author={Yang, Yujie and Jiang, Yuxuan and Liu, Yichen and Chen, Jianyu and Li, Shengbo Eben},
  journal={IEEE Robotics and Automation Letters},
  volume={8},
  number={3},
  pages={1295--1302},
  year={2023},
  publisher={IEEE}
}

@inproceedings{akametalu2014reachability,
  title={Reachability-based safe learning with Gaussian processes},
  author={Akametalu, Anayo K and Fisac, Jaime F and Gillula, Jeremy H and Kaynama, Shahab and Zeilinger, Melanie N and Tomlin, Claire J},
  booktitle={53rd IEEE conference on decision and control},
  pages={1424--1431},
  year={2014},
  organization={IEEE}
}

@article{fisac2018general,
  title={A general safety framework for learning-based control in uncertain robotic systems},
  author={Fisac, Jaime F and Akametalu, Anayo K and Zeilinger, Melanie N and Kaynama, Shahab and Gillula, Jeremy and Tomlin, Claire J},
  journal={IEEE Transactions on Automatic Control},
  volume={64},
  number={7},
  pages={2737--2752},
  year={2018},
  publisher={IEEE}
}

@article{shao2021reachability,
  title={Reachability-based trajectory safeguard (RTS): A safe and fast reinforcement learning safety layer for continuous control},
  author={Shao, Yifei Simon and Chen, Chao and Kousik, Shreyas and Vasudevan, Ram},
  journal={IEEE Robotics and Automation Letters},
  volume={6},
  number={2},
  pages={3663--3670},
  year={2021},
  publisher={IEEE}
}

@inproceedings{prajapat2024towards,
  title={Towards safe and tractable Gaussian process-based MPC: Efficient sampling within a sequential quadratic programming framework},
  author={Prajapat, Manish and Lahr, Amon and K{\"o}hler, Johannes and Krause, Andreas and Zeilinger, Melanie N},
  booktitle={2024 IEEE 63rd Conference on Decision and Control (CDC)},
  pages={7458--7465},
  year={2024},
  organization={IEEE}
}

@article{prajapat2025safe,
  title={Safe guaranteed exploration for non-linear systems},
  author={Prajapat, Manish and K{\"o}hler, Johannes and Turchetta, Matteo and Krause, Andreas and Zeilinger, Melanie N},
  journal={IEEE Transactions on Automatic Control},
  year={2025},
  publisher={IEEE}
}

@inproceedings{berkenkamp2017safe,
author = {Berkenkamp, Felix and Turchetta, Matteo and Schoellig, Angela P and Krause, Andreas},
booktitle = {Conference on Neural Information Processing Systems (NIPS)},
title = {{Safe Model-based Reinforcement Learning with Stability Guarantees}},
year = {2017}
}

@inproceedings{chow2018lyapunov,
    author = {Chow, Yinlam and Nachum, Ofir and Duenez-Guzman, Edgar and Ghavamzadeh, Mohammad},
    booktitle = {Conference on Neural Information Processing Systems (NIPS)},
    eprint = {1805.07708v1},
    title = {{A Lyapunov-based Approach to Safe Reinforcement Learning}},
    year = {2018}
}

@article{dalal2018safe,
  title={Safe exploration in continuous action spaces},
  author={Dalal, Gal and Dvijotham, Krishnamurthy and Vecerik, Matej and Hester, Todd and Paduraru, Cosmin and Tassa, Yuval},
  journal={arXiv preprint arXiv:1801.08757},
  year={2018}
}

@inproceedings{chow2019lyapunov,
    author = {Chow, Yinlam and Nachum, Ofir and Faust, Aleksandra and Duenez-Guzman, Edgar and Ghavamzadeh, Mohammad},
    booktitle = { RL4RealLife Workshop in the 36 th International Conference on Machine Learning},
    eprint = {1901.10031v2},
    title = {{Lyapunov-based Safe Policy Optimization for Continuous Control}},
    year = {2019}
}

@article{hewing2020learning,
  title={Learning-based model predictive control: Toward safe learning in control},
  author={Hewing, Lukas and Wabersich, Kim P and Menner, Marcel and Zeilinger, Melanie N},
  journal={Annual Review of Control, Robotics, and Autonomous Systems},
  volume={3},
  number={1},
  pages={269--296},
  year={2020},
  publisher={Annual Reviews}
}

@inproceedings{nguyen2024gameplay,
    title={Gameplay Filters: Robust Zero-Shot Safety through Adversarial Imagination},
    author={Duy Phuong Nguyen and Kai-Chieh Hsu and Wenhao Yu and Jie Tan and Jaime Fern{\'a}ndez Fisac},
    booktitle={8th Annual Conference on Robot Learning},
    year={2024}
}

@inproceedings{wendl2026safe,
    title={Safe Exploration via Policy Priors},
    author={Manuel Wendl and Yarden As and Manish Prajapat and Anton Pollak and Stelian Coros and Andreas Krause},
    booktitle={The Fourteenth International Conference on Learning Representations},
    year={2026}
}

@inproceedings{haarnoja2018soft,
  title={Soft actor-critic: Off-policy maximum entropy deep reinforcement learning with a stochastic actor},
  author={Haarnoja, Tuomas and Zhou, Aurick and Abbeel, Pieter and Levine, Sergey},
  booktitle={International conference on machine learning},
  pages={1861--1870},
  year={2018},
  organization={Pmlr}
}

@article{eisenberg1979proof,
  title={A proof of the hairy ball theorem},
  author={Eisenberg, Murray and Guy, Robert},
  journal={The American Mathematical Monthly},
  volume={86},
  number={7},
  pages={571--574},
  year={1979},
  publisher={Taylor \& Francis}
}

@article{gu2024review,
  title={A review of safe reinforcement learning: Methods, theories, and applications},
  author={Gu, Shangding and Yang, Long and Du, Yali and Chen, Guang and Walter, Florian and Wang, Jun and Knoll, Alois},
  journal={IEEE Transactions on Pattern Analysis and Machine Intelligence},
  volume={46},
  number={12},
  pages={11216--11235},
  year={2024},
  publisher={IEEE}
}

@article{garcia2012safe,
  title={Safe exploration of state and action spaces in reinforcement learning},
  author={Garcia, Javier and Fern{\'a}ndez, Fernando},
  journal={Journal of Artificial Intelligence Research},
  volume={45},
  pages={515--564},
  year={2012}
}

\end{document}